\documentclass[11pt,,screen,leftjustify]{acmart}

\makeatletter 
\usepackage{times}  
\usepackage{helvet} 
\usepackage{courier}  
\usepackage{graphicx} 
\usepackage{natbib}  
\usepackage{caption} 
\usepackage{booktabs}
\usepackage{url}
\usepackage[many]{tcolorbox}
\usepackage{amsmath, graphicx}
\usepackage{paralist} 
\usepackage{color}
\usepackage{multirow}
\usepackage{rotating}
\usepackage{todonotes}
\usepackage{amsbsy} 
\usepackage{subfigure}
\usepackage{booktabs}
\usepackage{xcolor}
\usepackage{tikz}
\usepackage{multirow}
\usepackage[mathscr]{eucal} 
\usepackage{epstopdf}
\usepackage{balance}
\usepackage{empheq} 

\usepackage{xcolor}
\usepackage{pifont}
\usepackage{xspace}
\usepackage{lscape}
\usepackage{rotating}

\newcommand{\cmark}{\color{teal}\ding{51}}%
\newcommand{\xmark}{\color{purple}\ding{55}}%

\newcommand{\Fakeddit}{\texttt{Fakeddit}\xspace}
\newcommand{\NewsBag}{\texttt{NewsBag}\xspace}
\newcommand{\NtfNews}{\texttt{N24News}\xspace}
\newcommand{\MMCOVID}{\texttt{MM-COVID}\xspace}
\newcommand{\ReCOVery}{\texttt{ReCOVery}\xspace}
\newcommand{\MMCoVaR}{\texttt{MMCoVaR}\xspace}
\newcommand{\IVC}{\texttt{Image-Verification-Corpus}\xspace}
\newcommand{\CoAID}{\texttt{CoAID}\xspace}
\newcommand{\MuMiN}{\texttt{MuMiN}\xspace}

\definecolor{lavender}{RGB}{230, 230, 250}
\definecolor{thistle}{RGB}{216, 191, 216}
\definecolor{plum}{RGB}{221, 160, 221}
\definecolor{orchid}{RGB}{218, 112, 214}
\definecolor{violet}{RGB}{238, 130, 238}
\definecolor{mediumpurple}{RGB}{147, 112, 219}
\definecolor{darkorchid}{RGB}{153, 50, 204}
\definecolor{darkviolet}{RGB}{148, 0, 211}
\definecolor{darkmagenta}{RGB}{139, 0, 139}
\definecolor{purple}{RGB}{128, 0, 128}
\usepackage{tikz}
\usetikzlibrary{shadows}
\tikzset{
  basic/.style  = {draw, rounded corners=6pt, drop shadow,text width=5cm, font=\sffamily, rectangle},
  root/.style   = {basic, rounded corners=6pt, thin, align=center,drop shadow,
                   fill=blue!20},
  level 2 red/.style = {basic, rounded corners=6pt, thin,align=center, fill=mediumpurple!50,drop shadow,
                   text width=10em, text height=0.8em},
level 2 green/.style = {basic, rounded corners=6pt, thin,align=center, fill=teal!50,
                   text width=9em,drop shadow,},
  level 3 red/.style = {basic,rounded corners=6pt, thin, align=center, fill=darkmagenta!15,
                 text width=11em,drop shadow,},
  level 3 green/.style = {basic,rounded corners=6pt, thin, align=center, fill=darkmagenta!15,
                 text width=11em,drop shadow,},
  level 4 red/.style = {basic,rounded corners=6pt, thin, align=center, fill=pink!60,
                 text width=11em,drop shadow,},
  level 4 green/.style = {basic,rounded corners=6pt, thin, align=center, fill=green!10,
                 text width=11em,drop shadow,}
}

\begin{document}

\title{Multi-modal Misinformation Detection: Approaches, Challenges and Opportunities}

\author{Sara Abdali}
\authornote{This material is based upon work supported by the National Science Foundation under Grant \#: 2127309 to the Computing Research Associate for the CIFellows Project. Any opinions, findings, and conclusions or recommendations expressed in this material are those of the author and do not necessarily reflect the views of funding agencies. The work has been done while the first author was a postdoctoral CIFellow at Georgia Tech. The first author is currently a  senior researcher at Microsoft Corp.
}
\email{sabdali3@gatech.edu}
\affiliation{%
  \institution{Georgia Institute of Technology }
  \streetaddress{School of Computational Social Science \& Eng., CODA building,756 W Peachtree St NW }
  \city{Atlanta}
  \state{Georgia}
  \country{USA}
  \postcode{30308}
}

\author{Sina Shaham}
\email{sshaham@usc.edu
}
\affiliation{%
  \institution{University of Southern California}
  \streetaddress{}
  \city{Los Angeles}
  \state{California}
  \country{USA}
  \postcode{}
}

\author{Bhaskar Krishnamachari}
\email{bkrishna@usc.edu}
\affiliation{%
  \institution{University of Southern California}
  \streetaddress{}
  \city{Los Angeles}
  \state{California}
  \country{USA}
  \postcode{}
}

\renewcommand{\shortauthors}{Abdali et al.}

\begin{abstract}
As social media platforms evolve from text-based forums into multi-modal environments, the nature of misinformation in social media is also transforming accordingly. Taking advantage of the fact that visual modalities such as images and videos are more favorable and attractive to users, and textual content is sometimes skimmed carelessly, misinformation spreaders have recently targeted contextual connections between the modalities, e.g., text and image. Hence, many researchers have developed automatic techniques for detecting possible cross-modal discordance in web-based content. We analyze, categorize, and identify existing approaches in addition to the challenges and shortcomings they face in order to unearth new research opportunities in the field of multi-modal misinformation detection.
\end{abstract}
\maketitle
\section{Introduction}
 Nowadays, billions of multi-modal posts containing texts, images, videos, soundtracks, etc., are shared throughout the web, mainly via social media platforms such as Facebook, Twitter, Snapchat, Reddit, Instagram, YouTube, and so on. While the combination of modalities allows for more expressive, detailed, and user-friendly content, it brings about new challenges, as it is harder to accommodate uni-modal solutions to multi-modal environments.
 \par However, in recent years, due to the sheer use of multi-modal platforms, many automated techniques for multi-modal tasks such as Visual Question Answering (VQA)~\cite{VQA_2017_Agrawal,Goyal2017MakingTV,VizWiz_2018_Gurari,GQA_Hudson_2019,Singh_VQAModels_2019}, image captioning~\cite{Captions_coco_chen_2015,Gurari_blindcaptioning_2020,Visual_Genome_2017_Krishna,young-etal-2014-image} and more recently for fake news detection including hate speech in multi-modal memes~\cite{Singhal2020SpotFakeAM,MM'21_Qi_Fuse_Diverse_Multimodal,DSAA'20_Giachanou_Multi-image,FAIR_mems} have been introduced by machine learning researchers.
 \par Similar to other multi-modal tasks, it is harder and more challenging to detect fake news on multi-modal platforms, as it requires not only the evaluation of each modality, but also cross-modal connections and credibility of the combination as well. This becomes even more challenging when each modality e.g., text or image is credible but the combination creates misinformative content. For instance, a COVID-19 anti-vaccination misinformation\footnote{``Misinformation'' is false information that spreads unintentionally, whereas the term ``Disinformation'' refers to false information that malicious users share intentionally and often strategically to
affect other audiences’ behaviors toward social, political, and economic events. In this work, regardless of spreaders’ intention, we refer to all sorts of false news i.e., misinformation and disinformation as ``Misinformation'' or ``Fake News'' interchangeably.} post can have text that reads ``vaccines do this'' and then attaches a graphic image of a dead person. In this case, although the image and text are not individually misinformative, taken together they create misinformation.
 \par Over the past decade, several detection models~\cite{shu2017fake,shu_weak_2020,Islam_2020_DeepLearning,cai_2020_A_Technical_Survey} have been developed to detect misinformation. However, the majority of them leverage only a single modality for misinformation detection, e.g., text~\cite{Horne:2017,wu2017gleaning,ASONAM2018,Weak_Supervision} or image~\cite{Huh_2018_Fighting_fake_news,ICDM_Qi_multidomain,VizFake,Choudhary_2021_ICSC}, which miss the important information conveyed by other modalities. There are existing works~\cite{Beyond,dEFEND,KNH,Abdali2020HiJoDSM,HAKAK_2021_ensemble} that leverage ensemble methods which create multiple models for each modality and then combine them to produce improved results. However, in many cases of multi-modal misinformation, loosely combining individual modalities is inadequate for detecting fake news, leading to the failure of the joint model.
\par Nevertheless, in recent years, machine learning scientists have developed different techniques for cross-modal fake news detection, which combine information from multiple modalities, leveraging cross-modal information such as the consistency and meaningful relationships between different modalities. Studying and analyzing these techniques and identifying existing challenges will give a clearer picture of the state of knowledge on multi-modal misinformation detection and open the door to new opportunities in this field.
\par Even though there are a number of valuable surveys on fake news detection~\cite{shu2017fake,kumar2018false,CardosoDurierdaSilva2019CanML}, very few of them focus on multi-modal techniques~\cite{multimodal_survey,alonsobartolome2021multimodal}. Since the number of proposed techniques for multi-modal fake news detection has been increasing immensely, the necessity of a comprehensive survey on existing techniques, datasets, and emerging challenges is felt more than ever. With that said, in this work, we aim to conduct a comprehensive study on fake news detection in multi-modal environments.
\par To this end, we classify multi-modal misinformation detection study into following directions:
\begin{itemize}
    \item \textbf{Multi-modal Data Study:} in this direction, the goal is to collect multi-modal fake news data, e.g., image, text, social context, etc., from different sources of information and use fact-checking resources to evaluate the veracity of the collected data and annotate them accordingly. Comparison and analysis of existing datasets, as well as benchmarking, are other tasks that fall under this category.
    \item \textbf{Multi-modal Feature Study:} the primary goal  of this study is to uncover significant links between various data modalities, which are frequently exploited by misinformation spreaders to distort, impersonate, or exaggerate original information. These meaningful connections may be used as clues for detecting misinformation in multi-modal environments such as social media posts. Another goal of this direction is to study and develop strategies for fusing features of different modalities and creating information-rich multi-modal features.
    \item \textbf{Multi-modal Model Study:} the main focus of this direction is on the development of efficient multi-modal machine learning solutions to detect misinformation by leveraging multi-modal features and clues. Proposing new techniques and approaches, in addition to improving the performance, scalability, interpretability, and explicability of machine learning models, are some of the common tasks in this direction.
\end{itemize}
\par
\begin{figure*}[!th]
    \begin{center}
  \includegraphics[width = 0.95\linewidth]{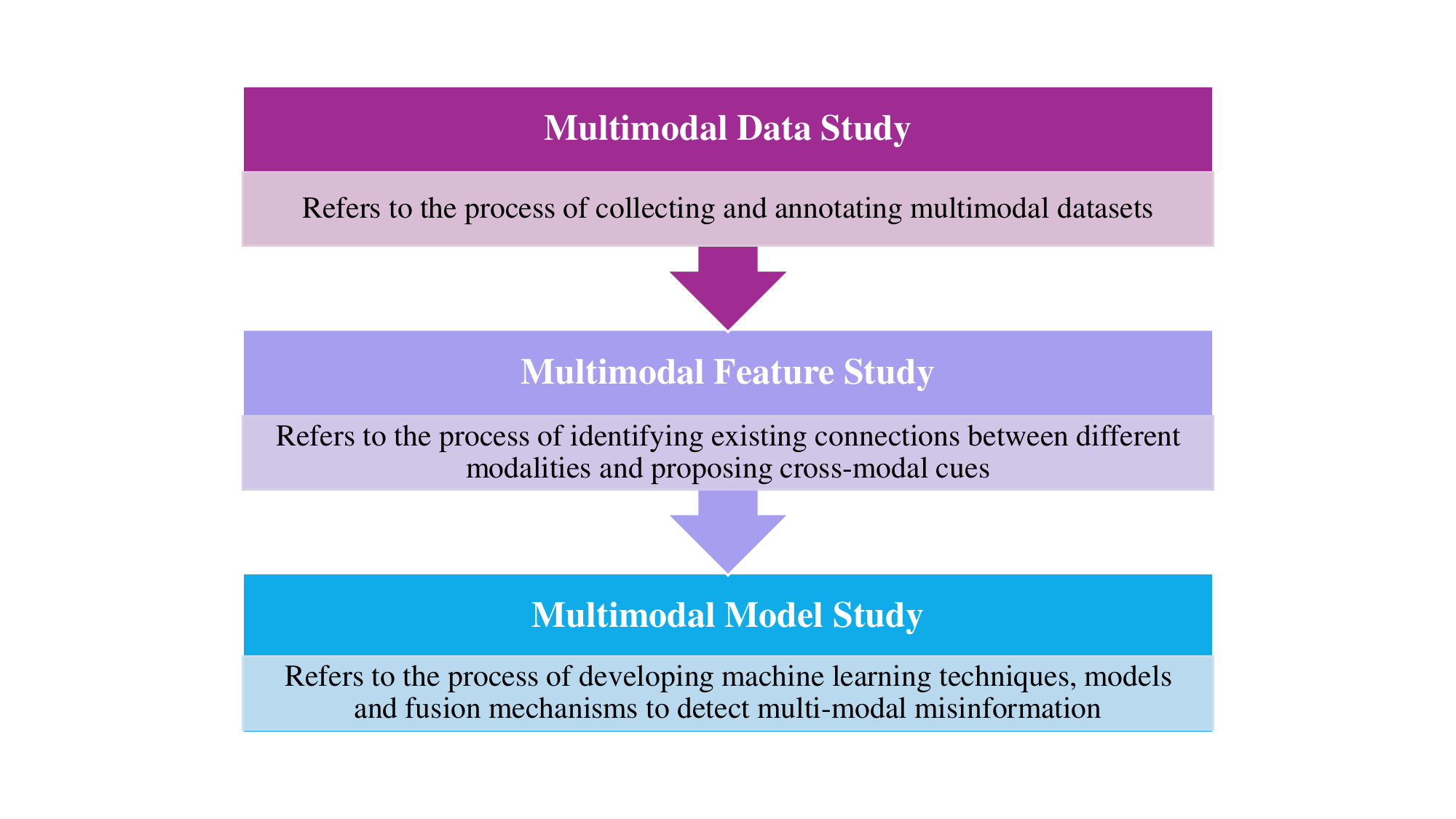}
    \end{center}
    \caption{An overview of multi-modal misinformation detection pipeline.}
        \label{fig:study}
\end{figure*}

 These three studies form a sequential pipeline in the multi-modal misinformation field, where the output of each study serves as the input for the next. Fig.~\ref{fig:study} provides a summary of these directions. In this work, we aim to explore each direction in greater depth to identify the challenges and shortcomings of each study and propose new avenues for addressing them.
\par The rest of this survey is organized as follows: In Section~\ref{Section: Multi-modal feature study}, we discuss the multi-modal feature study by introducing some widely spread categories of misinformation in multi-modal settings and commonly used cross-modal clues for detecting them. In the following section, we discuss different fusion mechanisms to merge modalities involved in such clues. Then, we explain the multi-modal model study by introducing solutions and categorizing them based on the machine learning techniques they utilize. In Section~\ref{Section: Multi-modal data study}, we describe the multi-modal data study by introducing, analyzing, and comparing existing databases for multi-modal fake news detection. In Section~\ref{sec:challenges}, we discuss existing challenges and shortcomings that each direction is facing. Finally, in Section~\ref{sec:opportunities}, we propose new avenues to address these shortcomings and advance multi-modal misinformation detection research.

\par We conducted our literature search across multiple databases, including IEEE Xplore, ACM Digital Library, and Google Scholar, using a combination of keywords related to our research focus. The inclusion criteria for the papers were defined by their relevance to the research question, publication date within the last ten years to ensure timeliness, and peer-reviewed status to guarantee quality. The selection process involved an initial screening of titles and abstracts, followed by a full-text review to confirm that each paper met our stringent criteria. This methodical approach ensures that the included papers provide a diverse yet focused perspective on the subject, offering readers a succinct and informative summary of current knowledge in the field. We emphasize the importance of transparency in our literature selection process and outline these steps to clarify the criteria and rationale behind our choices.

 \section{Multi-modal feature study}\label{Section: Multi-modal feature study}
 
In this section, we discuss the feature-based direction of multi-modal misinformation studies. To better understand the rationale behind multi-modal features and clues, we start with a brief introduction to some of the common categories of misinformation that spread in multi-modal environments. Furthermore, we discuss some of the commonly used multi-modal features and clues, and then we talk about existing fusion mechanisms for combining data modality features. Finally, we discuss the pros and cons of each fusion mechanism.
 \subsection{Common Categories of Misinformation in Multi-modal Environments}
\par Multi-modal misinformation refers to a package of misleading information that includes multiple modalities such as images, text, videos, and so on. In multi-modal misinformation, not all modalities are necessarily false, but sometimes the connections between the modalities are manipulated to deceive the audience’s perception. In what follows, we briefly discuss some of the common categories of misinformation that are widely spread in multi-modal settings. It is worth mentioning that these categories of misinformation are common in both multi-modal and uni-modal environments. However, we provide examples of each category in multi-modal platforms as well.
\begin{itemize}
    \item \textbf{Satire or Parody:}
This category refers to content that conveys true information with a satirical tone or added information that makes it false. One of the well-known publishers of this category is The Onion website\footnote{https://www.theonion.com/}, which is a digital media organization that publishes satirical articles on a variety of international, national, and local news. A multi-modal example of this category is an image within a satirical news article that contains absurd or ridiculous content or is manipulated to create humorous critique~\cite{Li2020AMM,phdthesis}. In this case, the textual content may not necessarily be false, but when combined with an image, it creates misleading content.
\item \textbf{Fabricated Content:} This category of information is completely false and is generated to deceive the audience. The intention behind publishing fabricated content is usually to mislead people for political, social, or economic benefits. A multi-modal instance of this category is a news report that uses auxiliary images or videos that are either completely fake or belong to irrelevant events.
 \item \textbf{Imposter Content:} This category of misinformation takes advantage of established news agencies by publishing misleading content under their branding. Since audiences trust established agencies, they are less likely to doubt the validity of the content and consequently pay less attention to subtle clues. Imposter content may damage the reputation of agencies and undermine audience trust. An example of imposter content is a website that mimics the domain features of global news outlets, such as CNN~\footnote{https://www.cnn.com/world} and BBC~\footnote{https://www.bbc.com/}. To detect this category of misinformation, it is crucial to identify and pay attention to the subtle features of web publishers~\cite{Abdali2020HiJoDSM,VizFake}.
\item \textbf{Manipulated Content:} This category of misinformation is generated by editing valid information, usually in the form of images and videos, to deceive audiences. Deepfake videos are well-known examples of this category. Manipulated videos and images have been widely generated to support fabricated content~\cite{TOLOSANA2020131,ff++,abdali2021deepfakerepresentationmultilinearregression}.
\item \textbf{False Connection:} This is one of the most common types of misinformation in multi-modal environments. In this category, some modalities, such as captions or titles, do not support other modalities, such as text or video. False connections are designed to catch the audience’s attention with clickbait headlines or provocative images~\cite{nakamura-etal-2020-fakeddit,li2020mmcovid}.
\end{itemize}
\par The above categories are used to spread a variety of fake news content such as ``Junk Science"\footnote{The term ``Junk Science" refers to inaccurate information about scientific facts that is used to skew opinions or push a hidden agenda}, ``Propaganda"\footnote{Refers to biased information that is often generated to promote a political point of view. Propaganda ranges from completely false information to subtle manipulation.}, ``Conspiracy Theories"\footnote{Refers to rejecting a widely accepted explanation for an event and offering a secret plot instead.}, ``Hate Speech", ``Rumors", ``Bias" etc. In the next section, we introduce some of cross-modal clues for detecting them in multi-modal settings. 
 \subsection{Multi-modal Features and Clues}
As previously indicated, combining features such as text and images has recently been utilized to identify false information in multi-modal contexts. In this section, we provide a non-exhaustive list of frequently used cues that machine learning researchers have used to identify false information. We emphasize that even though there are numerous other multi-modal combinations, they have not yet been fully explored by researchers at the time of writing, and we merely enumerate those that are frequently used in the literature.
\paragraph{Image and text mismatch}
The combination of textual content and article images is one of the widely used sets of features for multi-modal fake news detection. The intuition behind this cue is that some fake news spreaders use tempting images, such as exaggerated, dramatic, or sarcastic graphics, which are far from the textual content to attract users’ attention. Since it is difficult to find both pertinent and pristine images to match these fictions, fake news generators sometimes use manipulated images to support non-factual scenarios. Researchers refer to this cue as the similarity relationship between text and image~\cite{Zhou2020SAFESM,DSAA'20_Giachanou_Multi-image,Xue2021DetectingFN}, which could be captured with a variety of similarity measuring techniques such as cosine similarity between the title and image tags embeddings~\cite{Zhou2020SAFESM,DSAA'20_Giachanou_Multi-image} or similarity measure architectures~\cite{Xue2021DetectingFN}.
\paragraph{Mismatch between video and descriptive writing style}
On video-based platforms such as YouTube~\footnote{https://www.youtube.com/} and TikTok~\footnote{https://www.tiktok.com/}, video content is accompanied by descriptive textual information such as video descriptions, titles, users’ comments, and replies. Different users and video producers use various writing styles in such textual content. These writing styles can be learned and distinguished from unrecognized patterns by machine learning models. Meanwhile, the meaningful relationship between the visual content and the descriptive information, such as the video title, is another important clue that could be used for detecting online misbehavior~\cite{PRL'22_CHOI}. However, this is a very challenging task, as it is difficult to detect frames that are relevant to the text and discard irrelevant ones, such as advertisements, opening, or ending frames. Moreover, encoding all video frames is very inefficient in terms of speed and memory.
\paragraph{Textual content and propagation network}
The majority of online fact checkers, such as BS Detector\footnote{https://github.com/selfagency/bs-detector} or News Guard\footnote{https://www.newsguardtech.com/}, provide labels that pertain to domains rather than articles. Despite this disparity, several works~\cite{helmstetter2018weakly,brief_weakly} show that the weakly-supervised task of using labels pertaining to domains and subsequently testing on labels pertaining to articles yields negligible accuracy loss due to the strong correlation between the two~\cite{helmstetter2018weakly,brief_weakly}. Thus, by recognizing the domain features and behaviors, we might be able to classify articles published by them with admissible accuracy. Some of these feature patterns are the propagation network and word usage patterns of the domains, which could be considered~\cite{cross-domain,shu2019hierarchical,SILVA2021102618,Zhou_2019_networkbased} as a discriminating signature for different domains. It has been empirically shown that not only do news articles from different domains have significantly different word usage, but they also follow different propagation patterns~\cite{cross-domain}.
\paragraph{Textual content and overall look of serving domain}
Another domain-level feature that researchers have recently introduced for detecting misinformation is the overall look of the serving webpage~\cite{Abdali2020HiJoDSM,VizFake}. It is shown that, in contrast to credible domains, unreliable web-based news outlets tend to be visually busy and full of events such as advertisements, popups, etc.~\cite{VizFake}. Trustworthy webpages often look professional and ordered, as they often request users to agree to sign up or subscribe, have some featured articles, a headline picture, standard writing styles, and so on. On the other hand, unreliable domains tend to have an unprofessional blog-post style, negative space, and sometimes hard-to-read font errors. Considering this discriminating clue, researchers have recently proposed to consider the overall look of the webpages in addition to textual content and social context in order to create a multi-modal model for detecting misinformation~\cite{Abdali2020HiJoDSM,KNH}.
\paragraph{Video and audio mismatch}
Due to the ubiquity of camera devices and video-editing applications, video-based frameworks are extremely vulnerable to manipulation, e.g., virtual backgrounds, anime filters, etc. Such visual manipulations introduce non-trivial noise to the video frames, which may lead to the misclassification of irrelevant information from videos~\cite{BigData'21_Shang_TikTok}. Moreover, manipulated videos often incorporate content in different modalities such as audio and text, which sometimes are not misinformative when considered individually. However, they mislead the audience when considered jointly with the video content. To detect misleading content that is jointly expressed in video, audio, and text content, researchers have proposed leveraging frame-based information along with audio and text content on video-based platforms like TikTok~\cite{BigData'21_Shang_TikTok}.

\section{Multi-modal model study}\label{Section: Multi-modal model study}
Extracted features and the way they are fused play an important role in the model architecture. In fact, model-based and feature-based studies are closely related through fusion strategies, which makes the demarcation of these two studies very difficult. Hence, in this section, we first discuss common fusion strategies as the point of connection between the two studies. Furthermore, we categorize existing works based on the machine learning techniques exploited by each work. Specifically, we classify them into two main categories: 1) classic machine learning and 2) deep learning-based solutions. In this section, we discuss each category in detail.

\begin{figure*}[!hbt]
    \centering

\begin{subfigure}
    \centering
    \resizebox{0.85\textwidth}{!}{
    \begin{tikzpicture}

\node[draw, rectangle, fill=purple!40, drop shadow,minimum width=2cm, minimum height=2cm, align=center, rounded corners=6pt] (rect) at (0,0) {Early (Feature)\\ Fusion};

\node[draw, circle, fill=yellow!10,drop shadow, minimum size=1.5cm, align=center] (circ) at (6.4,0) {Classifier};

\draw[->, line width=0.3mm,>=stealth] (-4.5, 2) -- (rect.west) node[midway, above, sloped] {Textual Features};
\draw[->, line width=0.3mm,>=stealth] (-4.5, 0) -- (rect.west) node[near start, above, sloped] {Visual Features};
\draw[dotted,line width=0.4mm,>=stealth] (-5, -1.1) -- (rect.west);
\draw[-> ,line width=0.3mm,>=stealth] (-4.5, -2.8) -- (rect.west) node[midway, above, sloped] {Acoustic Features};

\draw[-> ,line width=0.3mm,>=stealth] (rect.east) -- (circ.west) node[midway, above] {mult-imodal Features};

\draw[-> ,line width=0.3mm,>=stealth] (circ.east) -- (9.7,0) node[ at end, above, sloped, rotate=270] {Final Decision};

        \end{tikzpicture}
        }
        \caption{Early fusion mechanism}
        \label{fig:subfig1}
    \end{subfigure}
    \hfill

    \begin{subfigure}
        \centering
        \resizebox{0.85\textwidth}{!}{
        \begin{tikzpicture}

        \node[draw, circle, fill=yellow!10, minimum size=1cm, align=center,drop shadow,] (circ1) at (0,4) {\small Classifier 1};
        \node[draw, circle, fill=yellow!10, minimum size=1cm, align=center,drop shadow,] (circ2) at (0,1.8) { \small Classifier 2};
        \node[draw, circle, fill=yellow!10, minimum size=2cm, align=center,drop shadow,] (circ3) at (0,-0.3) {\Large ...};
        \node[draw, circle, fill=yellow!10, minimum size=1cm, align=center,drop shadow,] (circ4) at (0,-2.4) {\small Classifier n};

        \node[draw, rectangle, fill=mediumpurple!30, minimum width=2cm, minimum height=2cm, align=center, rounded corners=6pt,drop shadow,] (rect) at (6,1) {Late (Decision)\\ Fusion};

        \draw[->, line width=0.3mm, >=stealth] (-5, 4) -- (circ1.west) node[midway, above, sloped] {Textual Features};
        \draw[->, line width=0.3mm, >=stealth] (-5, 1.8) -- (circ2.west) node[midway, above, sloped] {Visual Features};
        \draw[dotted,line width=0.4mm] (-5, -0.3) -- (circ3.west) node[] {};
        \draw[->, line width=0.3mm, >=stealth] (-5, -2.4) -- (circ4.west) node[midway, above, sloped] {Acoustic Features};

        \draw[->, line width=0.3mm, >=stealth] (circ1.east) -- (rect.west) node[midway, above, sloped] {Decision 1};
        \draw[->, line width=0.3mm, >=stealth] (circ2.east) -- (rect.west) node[midway, above, sloped] {Decision 2};
        \draw[dotted, line width=0.3mm, >=stealth] (circ3.east) -- (rect.west) node[midway, above, sloped] {...};
        \draw[->, line width=0.3mm, >=stealth] (circ4.east) -- (rect.west) node[midway, above, sloped] {Decision n};

        \draw[->, line width=0.3mm, >=stealth] (rect.east) -- (10,1) node[ at end, above, sloped, rotate=270] {Final Decision};

        \end{tikzpicture}
        }
        \caption{Late fusion mechanism.}
        \label{fig:subfig2}
    \end{subfigure}
\end{figure*}

\begin{figure}[!hbt]
    \centering
    \resizebox{0.85\textwidth}{!}{
    \begin{tikzpicture}

\node[draw, rectangle, fill=purple!40, minimum width=1cm, minimum height=2cm, align=center, rounded corners=6pt,drop shadow,] (rect) at (-0.8,0) {Early (Feature)\\ Fusion};

\node[draw, circle, fill=yellow!10, minimum size=1cm, align=center,drop shadow,] (circ) at (5.5,0) {Classifier};

\draw[->, line width=0.3mm,>=stealth] (-5, 2) -- (rect.west) node[midway, above, sloped] {Textual Features};
\draw[->, line width=0.3mm,>=stealth] (-5, 0) -- (rect.west) node[near start, above, sloped] {Visual Features};
\draw[dotted,  line width=0.4mm,>=stealth] (-5, -1.1) -- (rect.west);
\draw[-> ,line width=0.3mm,>=stealth] (-5, -2.8) -- (rect.west) node[midway, below, sloped] {Acoustic Features};

\draw[-> ,line width=0.3mm,>=stealth] (rect.east) -- (circ.west) node[midway, above] {mult-imodal Features};

\draw[-> ,line width=0.3mm,>=stealth] (circ.east) -- (8.3,0) node[midway, above] {Decision A};
    \node[draw, circle, fill=yellow!10, minimum size=1cm, align=center,drop shadow,] (circ1) at (-1,-4) {\small Classifier 1};
    \node[draw, circle, fill=yellow!10, minimum size=1cm, align=center,drop shadow,] (circ2) at (-1,-6.2) { \small Classifier 2};
    \node[draw, circle, fill=yellow!10, minimum size=2cm, align=center,drop shadow,] (circ3) at (-1,-8.3) { \Large ...};
    \node[draw, circle, fill=yellow!10, minimum size=1cm, align=center,drop shadow,] (circ4) at (-1,-10.4) { \small Classifier n};

    \node[draw, rectangle, fill=mediumpurple!30, minimum size=0.25cm,minimum height=9cm,
    align=center, rounded corners=6pt,drop shadow,] (rect3) at (8.55,-3) {\rotatebox{270}{Late (Decision) Fusion}};
    
    \node[draw, rectangle, fill=mediumpurple!30, minimum width=2cm, minimum height=2cm, align=center, rounded corners=6pt,drop shadow,] (rect) at (4,-7) {Late (Decision) \\ Fusion};

    \draw[->, line width=0.3mm, >=stealth] (-5, -4) -- (circ1.west) node[midway, above, sloped] {Textual Features};
    \draw[->, line width=0.3mm, >=stealth] (-5, -6.2) -- (circ2.west) node[midway, above, sloped] {Visual Features};
    \draw[dotted,line width=0.4mm] (-5, -8.3) -- (circ3.west) node[] {};
    \draw[->, line width=0.3mm, >=stealth] (-5, -10.4) -- (circ4.west) node[midway, above, sloped] {Acoustic Features};

    \draw[->, line width=0.3mm, >=stealth] (circ1.east) -- (rect.west) node[midway, above, sloped] {Decision 1};
    \draw[->, line width=0.3mm, >=stealth] (circ2.east) -- (rect.west) node[midway, below, sloped] {Decision 2};
    \draw[dotted, line width=0.3mm, >=stealth] (circ3.east) -- (rect.west) node[midway, above, sloped] {...};
    \draw[->, line width=0.3mm, >=stealth] (circ4.east) -- (rect.west) node[midway, above, sloped] {Decision n};

    \draw[->, line width=0.3mm, >=stealth] (rect.east) -- (8.3,-7) node[midway, above] {Decision B};

    \draw[->, line width=0.3mm, >=stealth] (rect3.east) -- (9.1,-3) node[at end, above, rotate=270] {Final Decision};

    \end{tikzpicture}
    }
    \caption{A hybrid of early and late fusion mechanisms.}
    \label{fig:connectingfig}
\end{figure}

\subsection{Fusion Mechanisms}
Data fusion is the process of combining information from multiple modalities to take advantage of all different aspects of the data and extract as much information as possible to improve the performance of machine learning models, as opposed to using a single data aspect or modality. Different fusion mechanisms have been used to combine features from different modalities, including those mentioned in the previous section. Fusion mechanisms are often categorized into one of the following groups:
\paragraph{Early fusion}
also known as feature-level fusion, this refers to combining features from different data modalities at an early stage using an operation, which is often concatenation. This type of fusion is often performed ahead of classification. If the fusion process is done after feature extraction, it is sometimes referred to as intermediate fusion~\cite{Deep_Multimodal_Fusion,Lahat2015MultimodalDF,Fusion_2021_ML}.  
\paragraph{Late fusion}
also known as decision-level or kernel-level fusion, this is usually done in the classification stage. This method depends on the results obtained by each data modality individually. In other words, the modality-wise classification results are combined using techniques such as sum, max, average, and weighted average. Most of the late fusion solutions use handcrafted rules, which are prone to human bias and are far from real-world peculiarities~\cite{Deep_Multimodal_Fusion,Lahat2015MultimodalDF,Fusion_2021_ML}.
\subsection{Comparison of fusion mechanisms} In most cases, early fusion is a complex operation, whereas late fusion is easier to perform~\cite{Atrey2010MultimodalFF} because, unlike early fusion where the features from different modalities (e.g., image and text) may have different representations, the decisions at the semantic level usually have the same representation. Therefore, the fusion of decisions is easier than the fusion of features. However, the late fusion strategy does not utilize the feature-level correlation among modalities, which may improve classification performance. In fact, it is shown that in many cases, the early fusion of different modalities outperforms multi-modal late fusion while applying deep learning or classic machine learning classifiers~\cite{fusion_evaluation,Gallo_IVCNZ_2020}. For instance, early fusion of images and texts while using BERT and CNN on the UPMC Food-101 dataset\footnote{http://visiir.lip6.fr/explore}~\cite{FoodDB_ICMEW_2015} outperforms late fusion of these modalities. 
\par Another advantage of early fusion is that it requires less computation time because training is performed only once, whereas late fusion needs multiple classifiers for local decisions~\cite{Atrey2010MultimodalFF}. However, to have the best of both worlds, there are hybrid approaches as well, which take advantage of both early and late fusion strategies~\cite{Atrey2010MultimodalFF}. Fig. \ref{fig:subfig1} to Fig. illustrate simplified schemes of various fusion mechanisms for multi-modal learning\ref{fig:connectingfig}. Traditional and modern approaches for detecting multi-modal misinformation, some of which employ fusion mechanisms, are covered in the section that follows.

\subsection{Classic Machine Learning Solutions}
As we discussed earlier, a vast majority of misinformation detection methods leverage a single modality, a.k.a. aspect of news articles, e.g., text~\cite{Horne:2017,wu2017gleaning,ASONAM2018,Weak_Supervision}, image~\cite{Huh_2018_Fighting_fake_news,ICDM_Qi_multidomain,VizFake,Choudhary_2021_ICSC}, user features~\cite{wu2018tracing,Shu_2018_Understanding_User,Shu_2019_user}, and temporal properties~\cite{kumar2018false,shu2019hierarchical,SONG2021102712}. However, recently, there have been very few works that incorporate various aspects of a news article using classic machine learning techniques to create multi-modal article representations.
\par For instance, a work by Shu et al.~\cite{Beyond} proposes individual embedded representations for text, user-user interactions, user-article interactions, and publisher-article interactions, and defines a joint optimization problem leveraging these individual representations. Finally, they apply a “Non-convex Optimization” solution via the Alternating Least Squares (ALS) algorithm to solve the proposed optimization problem.

\par In another work, Abdali et al.~\cite{Abdali2020HiJoDSM} propose an ``Algebraic Joint Structure'' algorithm called HiJoD, which encodes three different aspects of an article: the article text, the context of social sharing behaviors, and host website/domain features. These aspects are transformed into individual embeddings, and shared structures among these embeddings are extracted using a principled tensor-based framework. By canceling out the unshared structures, the extracted shared structures are then utilized for article classification. The classification performance of the algebraic joint model, HiJoD, is compared with the ``Naive Embeddings Concatenation'' of embedding representations. The results demonstrate that the tensor-based representation is more effective in capturing the nuanced patterns of the joint structure.
\par Another study~\cite{KNH} presents the K-Nearest Hyperplanes (KNH) graph, a new type of graph generalization where nodes are higher-order Euclidean subspaces formed by algebraic structures, aimed at multi-aspect modeling of news articles.
\par More recently, Meel et 
al.~\cite{HAKAK_2021_ensemble} have proposed an ``Ensemble Framework'' which leverages text embedding, a score calculated by cosine similarity between image caption and news body, and noisy images. Despite the fact that some of the modules of this model, e.g., text embedding generator, leverage deep attention-based architecture, the classification process is done via a classic ensemble technique, i.e., max ``Voting''. 

\par Summarily, due to the success of deep learning-based techniques in feature extraction and classification tasks, classic machine learning-based techniques are not commonly used these days. However, considering the fact that deep learning techniques are data-hungry and require a lot of effort for training and fine-tuning the models, depending on the applications, classic machine learning techniques are still being used solely or in conjunction with deep learning techniques. 

\subsection{Deep Learning Solutions}
Due to the impressive success of deep neural networks in feature extraction and classification of text, images, and many other modalities, they have been widely exploited by research scientists over the past few years for a variety of multi-modal tasks, including misinformation detection. We may categorize deep learning-based multi-modal misinformation detection into five categories: concatenation-based, attention-based, generative-based, graph neural network-based, and cross-modality discordance-aware architectures as demonstrated in Fig.~\ref{fig:categorization}. In what follows, we summarize and categorize the existing works into the aforementioned categories.

 \begin{figure*}[t!]
     \resizebox{0.8\textwidth}{!}{
\begin{tikzpicture}[
   level distance=1.5cm,
  level 1/.style={sibling distance=80mm},
  edge from parent/.style={->,draw},
  >=latex]

\node[root] {\textbf{\small Multi-modal \\Misinformation Detection}}
  child {node[level 2 red] (c1) {\scriptsize \textbf{Classic Machine Learning}}}
  child {node[level 2 red ] (c2) {\scriptsize \textbf{Deep Learning}}};

\begin{scope}[every node/.style={level 3 red}]

\scriptsize
\node [below of = c1,xshift=5pt] (c11) {Embeddings Concatenation};
\node [below of = c1,xshift=5pt,yshift=-30pt] (c12) {Algebraic \\Joint Structure};

\node [below of = c1,xshift=5pt,yshift=-60pt] (c13) {Ensemble \\Learning  \& Voting};

\node [below of = c1,xshift=5pt,yshift=-90pt] (c14) {Non-convex Optimization};

\node [below of = c2,xshift=5pt] (c21) {Concatenation-based Networks};
\node [below of = c2,xshift=5pt,yshift=-30pt] (c22) {Attention-based Networks};

\node [below of = c2,xshift=5pt,yshift=-60pt] (c23) {Graph Neural \\Networks};

\node [below of = c2,xshift=5pt,yshift=-95pt] (c24) {Cross-modal \\Discordance Aware Networks};

\node [below of = c2,xshift=5pt,yshift=-140pt] (c25) {Foundation Models \&  \\Prompt-based Techniques};
\end{scope}
\foreach \value in {1,2,3,4}
  \draw[->] (c1.180) |- (c1\value.west);

\foreach \value in {1,2,3,4,5}
  \draw[->] (c2.180) |- (c2\value.west);

\end{tikzpicture}
}
    \caption{ An overview of the multi-modal model study.}
        \label{fig:categorization}
\end{figure*}
\subsubsection{Concatenation-based Architectures}
The majority of the existing work on multi-modal misinformation detection embeds each modality, e.g., text or image, into a vector representation and then concatenates them to generate a multi-modal representation that can be utilized for classification tasks. For instance, Singhal et al. propose using pretrained XLNet and VGG-19 models to embed text and image, respectively, and then classify the concatenation of the resulting feature vectors to detect misinformation~\cite{Singhal2020SpotFakeAM}.
\par In another work~\cite{alonsobartolome2021multimodal}, Bartolome et al. exploit a Convolutional Neural Network (CNN) that takes as inputs both text and image corresponding to an article, and the outputs are concatenated into a single vector. Qi et al. extract text, Optical Character Recognition (OCR) content, news-related high-level semantics of images (e.g., celebrities and landmarks), and visual CNN features of the image. Then, in the stage of multi-modal feature fusion, text-image correlations, mutual enhancement, and entity inconsistency are merged by concatenation operation~\cite{MM'21_Qi_Fuse_Diverse_Multimodal}.
\par In another work~\cite{Rezayi_MIPR_2021}, Rezayi et al. leverage network, textual, and relaying features such as hashtags and URLs and classify articles using the concatenation of the feature embeddings. Works in~\cite{dEFEND, ARCNN_Raj_2022} are other examples of this category of deep learning-based solutions.

\subsubsection{Attention-based Architectures}
As mentioned above, many architectures simply concatenate vector representations, thereby failing to build effective multi-modal embeddings. Such models are not efficient in many cases. For instance, the entire text of an article does not necessarily need to be false for the corresponding image and vice versa to consider the article as misinformative content. Thus, some recent works attempt to use the attention mechanism to attend to relevant parts of images, text, etc. The attention mechanism is a more effective approach for utilizing embeddings, as it produces richer multi-modal representations.
\par For instance, a work by Sachan et al.~\cite{SCATE_ASONOM2021} proposes Shared Cross Attention Transformer Encoders (SCADE), which exploits CNNs and transformer-based methods to encode image and text information and utilizes cross-modal attention and shared layers for the two modalities. SCADE pays attention to the relevant parts of each modality with reference to the other.
\par Another example is a work by Kumari et al.~\cite{KUMARI2021115412}, where a framework is developed to maximize the correlation between textual and visual information. This framework has four different sub-modules: Attention-Based Stacked Bidirectional Long Short Term Memory (ABS-BiLSTM) for textual feature representation, Attention-Based Multilevel Convolutional Neural Network–Recurrent Neural Network (ABM-CNN–RNN) for visual feature extraction, multi-modal Factorized Bilinear Pooling (MFB) for feature fusion, and finally Multi-Layer Perceptron (MLP) for classification.
 
\par In another study, Qian et al.~\cite{Hierarchical_Contextual_Attention} introduce the Hierarchical Multi-modal Contextual Attention Network (HMCAN) architecture. This architecture leverages a pre-trained BERT and convolutional ResNet50 to create word and image embeddings. It also employs a multi-modal contextual attention network to investigate multi-modal context information. HMCAN uses various multi-modal contextual attention networks to form a hierarchical encoding network, aiming to explore and capture the rich hierarchical semantics of multi-modal data.
\par Another example is~\cite{Jin_2017_MM}, where Jin et al. fuse features from three modalities, i.e., textual, visual, and social context, using an RNN that utilizes an attention mechanism (att-RNN) for feature alignment. Jing et al. propose TRANSFAKE~\cite{TRANSFAKE_IJCNN_2021} to connect features of text and images into a series and feed them into a vision-language transformer model to learn the joint representation of multi-modal features. TRANSFAKE adopts a preprocessing method similar to BERT for concatenated text, comments, and images.
\par In another work~\cite{FMFN_2022AppliedSciences}, Wang et al. apply scaled dot-product attention on top of image and text features as a fine-grained fusion and use the fused feature to classify articles.
\par Wang et al. propose a deep learning network for biomedical informatics that leverages visual and textual information and a semantic- and task-level attention mechanism to focus on the essential contents of a post that signal anti-vaccine messages~\cite{Wang_2021_Biomedical}.
\par Another example is the study by Lu et al., where they concatenate representations of user interaction, word representations, and propagation features after implementing a dual co-attention mechanism. The goal is to capture the correlations between users’ interactions/propagation and the tweet’s text~\cite{lu-li-2020-gcan}.
\par Finally, Song et al.~\cite{Song2021AMF} propose a multi-modal fake news detection architecture based on Cross-modal Attention Residual (CARN) and Multichannel Convolutional Neural Networks (CARMN). Cross-modal Attention Residual (CARN) selectively extracts the information related to a target modality from a source modality while maintaining the unique information of the target.

\subsubsection{Generative Architectures}
In this category of deep learning solutions, the goal is to either apply Generative Networks or use auxiliary networks to learn individual or multi-modal representations, spaces, or parameters in order to improve the classification performance of the fake news detector.

\par As an example, Jaiswal et al. propose a BERT-based multi-modal variational Autoencoder (VAE)~\cite{UPCON_2021_Jaiswal} that consists of an encoder, decoder, and a fake news detector. The encoder encodes the shared representations of both the image and text into a multidimensional latent vector. The decoder decodes the multidimensional latent vector into the original image and text, and the fake news detector is a binary classifier that takes the shared representation as an input and classifies it as either fake or real.

\par Similarly, Kattar et al. propose a deep multi-modal variational autoencoder (MVAE)~\cite{Multimodal_Variational_Autoencoder2019} which learns a unified representation of both the modalities of a tweet’s content. Similar to the previous work, MVAE has three main components: an encoder, a decoder, and a fake news detector that utilizes the learned shared representation to predict if a news is fake or not.

\par Like the previous work, a work by Zeng et al.~\cite{Zeng_deep_correlations_2020} proposes to capture the correlations between text and image by a VAE-based multi-modal feature fusion method. In another work, Wang et al. propose Event Adversarial Neural Networks (EANN)~\cite{KDD'18_Wang_EANN}, an end-to-end framework which can derive event-invariant features and thus benefit the detection of fake news on newly arrived events. It consists of three main components: a multi-modal feature extractor, the fake news detector, and the event discriminator. The multi-modal feature extractor is responsible for extracting the textual and visual features from posts. It cooperates with the fake news detector to learn the discriminating representation of news articles. The role of the event discriminator is to remove the event-specific features and keep shared features among the events.

\par In another work~\cite{Wang2021MultimodalEF}, Wang et al. propose the MetaFEND framework, which is able to detect fake news on emergent events with a few verified posts using an event adaptation strategy. The MetaFEND framework has two stages: event adaptation and detection. In the event adaptation stage, the model adapts to specific events, and then in the detection stage, the event-specific parameter is leveraged to detect fake news on a given event. Although MetaFEND does not apply a generative architecture, it leverages an auxiliary network to learn an event-specific parameter set to improve the efficiency of the fake news detector.

\par The last example is a work~\cite{cross-domain} by Silva et al., where they propose a cross-domain framework using text and propagation network. The proposed model consists of two components: an unsupervised domain embedding learning and a supervised domain-agnostic news classification. The unsupervised domain embedding exploits text and propagation network to represent a news domain with a low-dimensional vector. The classification model represents each news record as a vector using the textual content and the propagation network. Then, the model maps this representation into two different subspaces such that one preserves the domain-specific information. Later on, these two components are integrated to identify fake news while exploiting domain-specific and cross-domain knowledge in the news records.

\subsubsection{Graph Neural Network Architectures}
In recent years, Graph Neural Networks (GNNs) have been successfully exploited for fake news detection~\cite{GNNfakenews1,GNNfakenews2,GNNfakenews3}, thereby catching researchers' attention for multi-modal misinformation detection tasks as well. In this category of deep learning solutions, article content (e.g., text, image, etc.) is represented by graphs, and then graph neural networks are used to extract the semantic-level features.

\par For instance, Wang et al. construct a graph for each social media post based on the point-wise mutual information (PMI) score of pairs of words, extracted objects in visual content, and knowledge concepts through knowledge distillation. They then utilize a Knowledge-driven Multi-modal Graph Convolutional Network (KMGCN) which extracts the multi-modal representation of each post through graph convolutional networks~\cite{KMGCN_Wang_2020}.

\par Another GCN-based model is GAME-ON~\cite{GAME-ON_Dhawan2022}, which represents each news item with uni-modal visual and textual graphs and then projects them into a common space. To capture multi-modal representations, GAME-ON applies a graph attention layer on a multi-modal graph generated out of modality graphs.

\subsubsection{Cross-modal Discordance-aware Architectures}
In the previously discussed categories, deep learning models are employed to merge different modalities to create distinguishing representations. However, in this category, deep learning architectures are tailored to address identified discrepancies between modalities. The idea is that fabricating either modality will cause dissonance between them, leading to misrepresented, misinterpreted, and misleading news. Therefore, subtle cross-modal discordance clues can be identified and learned by customized architectures. Consequently, methods utilizing ``contrastive learning'' or Contrastive Language-Image Pre-Training (CLIP) based architectures~\cite{10096771,Jiang2023SimilarityAwareMP} may fall into this category.

\par In many cases, fake news propagators use irrelevant modalities (e.g., image, video, audio, etc.) for false statements to attract readers’ attention. Thus, the similarity of text to other modalities (e.g., image, audio, etc.) is a cue for measuring the credibility of a news article.

\par With that said, Zhou et al.~\cite{Zhou2020SAFESM} propose SAFE, a Similarity-Aware Multi-Modal Fake News Detection framework by defining the relevance between news textual and visual information using a modified cosine similarity.

\par Similarly, Giachanou et al. propose a multi-image system that combines textual, visual, and semantic information~\cite{DSAA'20_Giachanou_Multi-image}. The semantic representation refers to the text-image similarity calculated using the cosine similarity between the title and image tag embeddings.

\par In another work, Singhal et al.~\cite{MMAsia'21_Singhal_Inter-Modality} develop an inter-modality discordance-based fake news detector which learns discriminating features and employs a modified version of contrastive loss that explores the inter-modality discordance.

\par Xue et al.~\cite{Xue2021DetectingFN} propose a Multi-modal Consistency Neural Network (MCNN) which utilizes a similarity measurement module that measures the similarity of multi-modal data to detect the possible mismatches between the image and text. Lastly, Biamby et al.~\cite{biamby2021twittercomms} leverage the CLIP model~\cite{Radford2021LearningTV} to jointly learn image/text representation to detect image-text inconsistencies in tweets. Instead of concatenating vector representations, CLIP jointly trains an image encoder and a text encoder to predict the correct pairings of a batch of (image, text) training examples.

\par On video-based platforms such as YouTube videos, typically different producers use different title and description, as users and subscribers express their opinions in different writing styles. \par Having this clue in mind, Choi et al. propose a framework to identify fake content on YouTube~\cite{PRL'22_CHOI}. They propose to use domain knowledge and “hit-likes”
of comments to create the comments embedding which is effective
in detecting fake news videos. They encode Multi-modal features i.e., image and text and detect differences between title, description or video and user’s comments.
\par In another work~\cite{BigData'21_Shang_TikTok}, Shang et al. develop TikTec, a
multi-modal misinformation detection framework that explicitly
exploits the captions to accurately capture the key information
from unreliable video content. This framework learns the composed misinformation that is jointly conveyed by the visual
and audio content. TikTec consists of four
major components. A Caption-guided Visual Representation Learning (CVRL) module which 
identify the misinformation-related visual features of each
sampled video frame, An Acoustic-aware Speech
Representation Learning (ASRL) module that
jointly learns the misleading semantic information that is
deeply embedded in the unstructured and casual audio tracks
and the Visual-speech Co-attentive Information Fusion (VCIF) module which captures the multiview composed information jointly embedded in the heterogeneous visual and audio contents of the video. Finally, the Supervised Misleading Video Detection (SMVD) module identifies misleading COVID-19 videos.

\subsubsection{Foundation Models and Prompt-based Techniques}
A foundation model is a large machine learning model that is trained on large-scale datasets such that it can be adapted to a wide range of downstream tasks. Some examples of multi-modal foundation models are pre-trained GPT-4~\cite{openai2023gpt4}, DALL-E~\cite{Ramesh2021ZeroShotTG}, Florence~\cite{Yuan2021FlorenceAN}, Flamingo~\cite{Alayrac2022FlamingoAV}, and so on.

In-Context Learning (ICL) is the simplest and one of the most effective ways of using foundation models. ICL is a training-free technique where models learn to learn from limited demonstrations and descriptions and generalize to unseen tasks~\cite{Tai2023LinkContextLF}. The learn-to-learn concept was first introduced in meta-learning, which is a family of machine learning techniques that uses few examples to adapt the model to new tasks. In recent years, meta-learning has been used for different applications, including multi-modal misinformation detection~\cite{Wang2021MultimodalEF,yue2023metaadapt}. However, the GPT-3 paper~\cite{Brown2020LanguageMA} shows that few-shot learning is an emergent capability of Large Language Models (LLMs) and could be taken advantage of using ICL techniques. In fact, a frozen model can be conditioned to perform a variety of tasks through ICL, where a user primes the model for a given task through prompt design, i.e., manually crafting a text prompt with descriptions or examples of the task.

\par A more effective way to condition frozen models is by using tunable prompts. Unlike model fine-tuning, which modifies the model’s parameters through additional training on new data, prompt-tuning adjusts the parameters of the prompt tokens while keeping the pre-trained model frozen~\cite{lester-etal-2021-power}.

\par ICL techniques, including few-shot and zero-shot prompting, as well as prompt tuning, have been widely used to query Large Language Models (LLMs) for a variety of downstream tasks, including misinformation detection. For example, Jiang et al.~\cite{jiang2022fake} study the role of prompt learning in detecting fake news. In another work~\cite{gao2023few}, Gao et al. put forward a prompt-tuning template to extract knowledge from a pretrained LM for detecting misinformation. Another example is a work by Tian et al.~\cite{tian2023metatroll}, where few-shot learning is leveraged for troll detection. In another work by Lin et al.~\cite{lin2023zero}, prompt tuning is used for rumor detection using a zero-shot framework. Similarly, ~\cite{zuo2022continually} presents a continual learning framework that applies prompt tuning for rumor detection.
\par However, there are few existing works that utilize them for misinformation detection in multi-modal settings. One of the existing works is PromptHate~\cite{Cao2023PromptingFM}, a simple prompt-based multi-modal model that prompts pre-trained language models (PLMs) for hateful meme classification.
PromptHate constructs simple prompts and provides a few in-context examples to exploit the implicit
knowledge in the pre-trained RoBERTa to classify hateful memes. 
\par In another work~\cite{10096771}, a novel
propaganda detection model, Antipersuasion Prompt Enhanced Contrastive Learning (APCL), is
proposed for detecting propaganda. The prompt is designed with a persuasion prompt template and an anti-persuasion prompt template to build matched text-image and mismatched text-image
pairs, respectively. Later on, the distances between the two prompt templates and pairs of text and
image are used for detection. 
\par More recently, Cao et al. leverage pre-trained vision-language models
(PVLMs) in a zero-shot and fine-tuning-free VQA setting to address the problem of meme detection
by generating hateful content-centric image captions~\cite{Cao2023ProCapLA}.
\par In addition, Jian et al. propose a Similarity-Aware Multimodal Prompt Learning (SAMPLE)
framework that incorporates prompt-tuning into multi-modal fake news detection~\cite{Jiang2023SimilarityAwareMP}. SAMPLE uses
three prompt templates: discrete prompting, continuous prompting, and mixed prompting to the
original input text, and employs the pre-trained RoBERTa to extract text features from the prompt.
Furthermore, the pre-trained CLIP is used to obtain the input texts, input images, and their semantic
similarities. SAMPLE introduces a similarity-aware multi-modal feature fusing approach that applies
standardization and a Sigmoid function to adjust the intensity of the final cross-modal representation
and mitigate noise injection via uncorrelated cross-modal features.

\section{Multi-modal Data Study}\label{Section: Multi-modal data study}
Data acquisition and preparation are the most important building blocks of a machine learning pipeline. Machine learning models leverage training data to continuously improve themselves over time. Thus, sufficient good quality, and in most cases annotated data, is extremely crucial for these models to operate effectively. With that said, in this section, we introduce and compare some of the existing multi-modal datasets for the fake news detection task. Later on, we will discuss some of the limitations of these datasets.\\

\textbf{\IVC}:\footnote{https://githubhelp.com/MKLab-ITI/image-verification-corpus} is an evolving dataset containing 17,806 fake and real posts with images shared on Twitter. This dataset is created as an open corpus of tweets containing images that may be used for assessing online image verification approaches (based on tweet texts and user features), as well as building classifiers for new content. Fake and real images in this dataset have been annotated by online sources that evaluate the credibility of the images and the events they are associated with~\cite{boididou2018detection}.\\

\begin{table*}[t]
\renewcommand{\arraystretch}{1.1}  
\centering
\scriptsize
\caption{A summary of the existing deep learning-based solutions.}
 \begin{tabular}{p{10mm}p{15mm}p{15mm}p{15mm}p{15mm}p{20mm}p{20mm}p{20mm}p{20mm}|}  
 \hline
\textbf{Paper}&\textbf{\parbox{50pt}{Concat.}}& \textbf{\parbox{50pt}{Attention }}&\textbf{\parbox{50pt}{Generative }}&\textbf{\parbox{50pt}{GNN}}&\textbf{{\parbox{50pt}{Cross-modal Cue}}}&\textbf{{\parbox{63pt}{Prompting}}}&\textbf{Primary Focus}\\  

\midrule
\cite{Singhal2020SpotFakeAM}&\texttt{\cmark} & \texttt{\xmark} & \texttt{\xmark} &\texttt{\xmark} & \texttt{\xmark} & \texttt{\xmark} & Concatenation\\
\cite{dEFEND}&\texttt{\cmark}&\texttt{\xmark}&\texttt{\xmark}&\texttt{\xmark}&\texttt{\xmark}&\texttt{\xmark} & Concatenation \\
\cite{alonsobartolome2021multimodal}&\texttt{\cmark}&\texttt{\xmark}&\texttt{\xmark}&\texttt{\xmark}&\texttt{\xmark}&\texttt{\xmark} & Concatenation\\ \cite{MM'21_Qi_Fuse_Diverse_Multimodal}&\texttt{\cmark}&\texttt{\xmark}&\texttt{\xmark}&\texttt{\xmark}&\texttt{\xmark}&\texttt{\xmark} & Concatenation \\
\cite{Rezayi_MIPR_2021}&\texttt{\cmark}&\texttt{\xmark}&\texttt{\xmark}&\texttt{\xmark}&\texttt{\xmark}&\texttt{\xmark} & Concatenation \\
\cite{ARCNN_Raj_2022}&\texttt{\cmark}&\texttt{\xmark}&\texttt{\xmark}&\texttt{\xmark}&\texttt{\xmark}&\texttt{\xmark} & Concatenation \\
\midrule
\cite{Jin_2017_MM}&\texttt{\cmark}&\texttt{\cmark}&\texttt{\xmark}&\texttt{\xmark}&\texttt{\xmark}&\texttt{\xmark} & Attention Mech.\\
\cite{lu-li-2020-gcan}&\texttt{\cmark}&\texttt{\cmark}&\texttt{\xmark}&\texttt{\xmark}&\texttt{\cmark}&\texttt{\xmark} & Attention Mech.\\
\cite{Hierarchical_Contextual_Attention}&\texttt{\cmark}&\texttt{\cmark}&\texttt{\xmark}&\texttt{\xmark}&\texttt{\xmark}&\texttt{\xmark} & Attention Mech.\\
\cite{messina-etal-2021-aimh}&\texttt{\cmark}&\texttt{\cmark}&\texttt{\xmark}&\texttt{\xmark}&\texttt{\xmark}&\texttt{\xmark} & Attention Mech.\\
\cite{SCATE_ASONOM2021}&\texttt{\cmark}&\texttt{\cmark}&\texttt{\xmark}&\texttt{\xmark}&\texttt{\xmark}&\texttt{\xmark} & Attention Mech.\\
\cite{KUMARI2021115412}&\texttt{\xmark}&\texttt{\cmark}&\texttt{\xmark}&\texttt{\xmark}&\texttt{\xmark}&\texttt{\xmark} & Attention Mech.\\
\cite{TRANSFAKE_IJCNN_2021}&\texttt{\cmark}&\texttt{\cmark}&\texttt{\xmark}&\texttt{\xmark}&\texttt{\xmark}&\texttt{\xmark} & Attention Mech.\\
\cite{Wang_2021_Biomedical}&\texttt{\cmark}&\texttt{\cmark}&\texttt{\xmark}&\texttt{\xmark}&\texttt{\xmark}&\texttt{\xmark} & Attention Mech.\\
\cite{Song2021AMF}&\texttt{\cmark}&\texttt{\cmark}&\texttt{\xmark}&\texttt{\xmark}&\texttt{\xmark}&\texttt{\xmark} & Attention Mech.\\
\cite{FMFN_2022AppliedSciences}&\texttt{\cmark}&\texttt{\cmark}&\texttt{\xmark}&\texttt{\xmark}&\texttt{\xmark}&\texttt{\xmark} & Attention Mech.\\
\cite{AENeT_Jain_2022}&\texttt{\cmark}&\texttt{\cmark}&\texttt{\xmark}&\texttt{\xmark}&\texttt{\xmark}&\texttt{\xmark} & Attention Mech.\\

\midrule
\cite{KDD'18_Wang_EANN}&\texttt{\cmark}&\texttt{\xmark}&\texttt{\cmark}&\texttt{\xmark}&\texttt{\xmark}&\texttt{\xmark} & Generative Net.\\
\cite{Multimodal_Variational_Autoencoder2019}&\texttt{\cmark}&\texttt{\xmark}&\texttt{\cmark}&\texttt{\xmark}&\texttt{\xmark}&\texttt{\xmark} & Generative Net.\\
\cite{Zeng_deep_correlations_2020}&\texttt{\cmark}&\texttt{\cmark}&\texttt{\cmark}&\texttt{\xmark}&\texttt{\xmark}&\texttt{\xmark} & Generative Net.\\
\cite{UPCON_2021_Jaiswal}&\texttt{\cmark}&\texttt{\cmark}&\texttt{\cmark}&\texttt{\xmark}&\texttt{\xmark}&\texttt{\xmark} & Generative Net.\\
\cite{cross-domain}&\texttt{\cmark}&\texttt{\xmark}&\texttt{\cmark}&\texttt{\xmark}&\texttt{\xmark}&\texttt{\xmark} & Generative Net.\\
\cite{ZHOU2022116517}&\texttt{\cmark}&\texttt{\xmark}&\texttt{\cmark}&\texttt{\xmark}&\texttt{\xmark}&\texttt{\xmark} & Generative Net.\\
\cite{Wang2021MultimodalEF}&\texttt{\cmark}&\texttt{\cmark}&\texttt{\cmark}&\texttt{\xmark}&\texttt{\xmark}&\texttt{\xmark} & Generative Net.\\
\midrule
\cite{KMGCN_Wang_2020}&\texttt{\xmark}&\texttt{\xmark}&\texttt{\xmark}&\texttt{\cmark}&\texttt{\xmark}&\texttt{\xmark} & GNN\\
\cite{GAME-ON_Dhawan2022}&\texttt{\xmark}&\texttt{\cmark}&\texttt{\xmark}&\texttt{\cmark}&\texttt{\xmark}&\texttt{\xmark} & GNN\\
\midrule

\cite{Zhou2020SAFESM}&\texttt{\xmark}&\texttt{\xmark}&\texttt{\xmark}&\texttt{\xmark}&\texttt{\cmark}&\texttt{\xmark} &  Cross-Modal Cue\\
\cite{DSAA'20_Giachanou_Multi-image}&\texttt{\cmark}&\texttt{\cmark}&\texttt{\xmark}&\texttt{\xmark}&\texttt{\cmark}&\texttt{\xmark} & Cross-Modal Cue\\
\cite{Xue2021DetectingFN}&\texttt{\xmark}&\texttt{\cmark}&\texttt{\xmark}&\texttt{\xmark}&\texttt{\cmark}&\texttt{\xmark} & Cross-Modal Cue\\
\cite{BigData'21_Shang_TikTok}&\texttt{\cmark}&\texttt{\cmark}&\texttt{\xmark}&\texttt{\xmark}&\texttt{\cmark}&\texttt{\xmark} &  Cross-Modal Cue\\
\cite{MMAsia'21_Singhal_Inter-Modality}&\texttt{\cmark}&\texttt{\cmark}&\texttt{\xmark}&\texttt{\xmark}&\texttt{\cmark}&\texttt{\xmark} &  Cross-Modal Cue\\
\cite{biamby2021twittercomms}&\texttt{\xmark}&\texttt{\xmark}&\texttt{\xmark}&\texttt{\xmark}&\texttt{\cmark}&\texttt{\xmark} &  Cross-Modal Cue\\
\cite{PRL'22_CHOI}&\texttt{\cmark}&\texttt{\cmark}&\texttt{\xmark}&\texttt{\xmark}&\texttt{\cmark}&\texttt{\xmark} & Cross-Modal Cue\\
\midrule

\cite{Cao2023PromptingFM}&\texttt{\xmark}&\texttt{\xmark}&\texttt{\xmark}&\texttt{\xmark}&\texttt{\cmark}&\texttt{\cmark} & Prompting\\
\cite{10096771}&\texttt{\xmark}&\texttt{\xmark}&\texttt{\xmark}&\texttt{\xmark}&\texttt{\cmark}&\texttt{\cmark} & Prompting\\
\cite{Cao2023ProCapLA}&\texttt{\xmark}&\texttt{\xmark}&\texttt{\xmark}&\texttt{\xmark}&\texttt{\cmark}&\texttt{\cmark} & Prompting\\
\cite{Jiang2023SimilarityAwareMP}&\texttt{\cmark}&\texttt{\xmark}&\texttt{\xmark}&\texttt{\xmark}&\texttt{\cmark}&\texttt{\cmark} & Prompting\\
\bottomrule
\end{tabular}
\label{table:summary}
 \end{table*}
\clearpage

\begin{figure}[!ht]
    \centering
    \subfigure[\textbf{ Caption:} These squirrels fighting. \newline \textbf{     Class: \color{teal}\textbf{True Content} } ]
    {
        \includegraphics[width=0.33\textwidth]{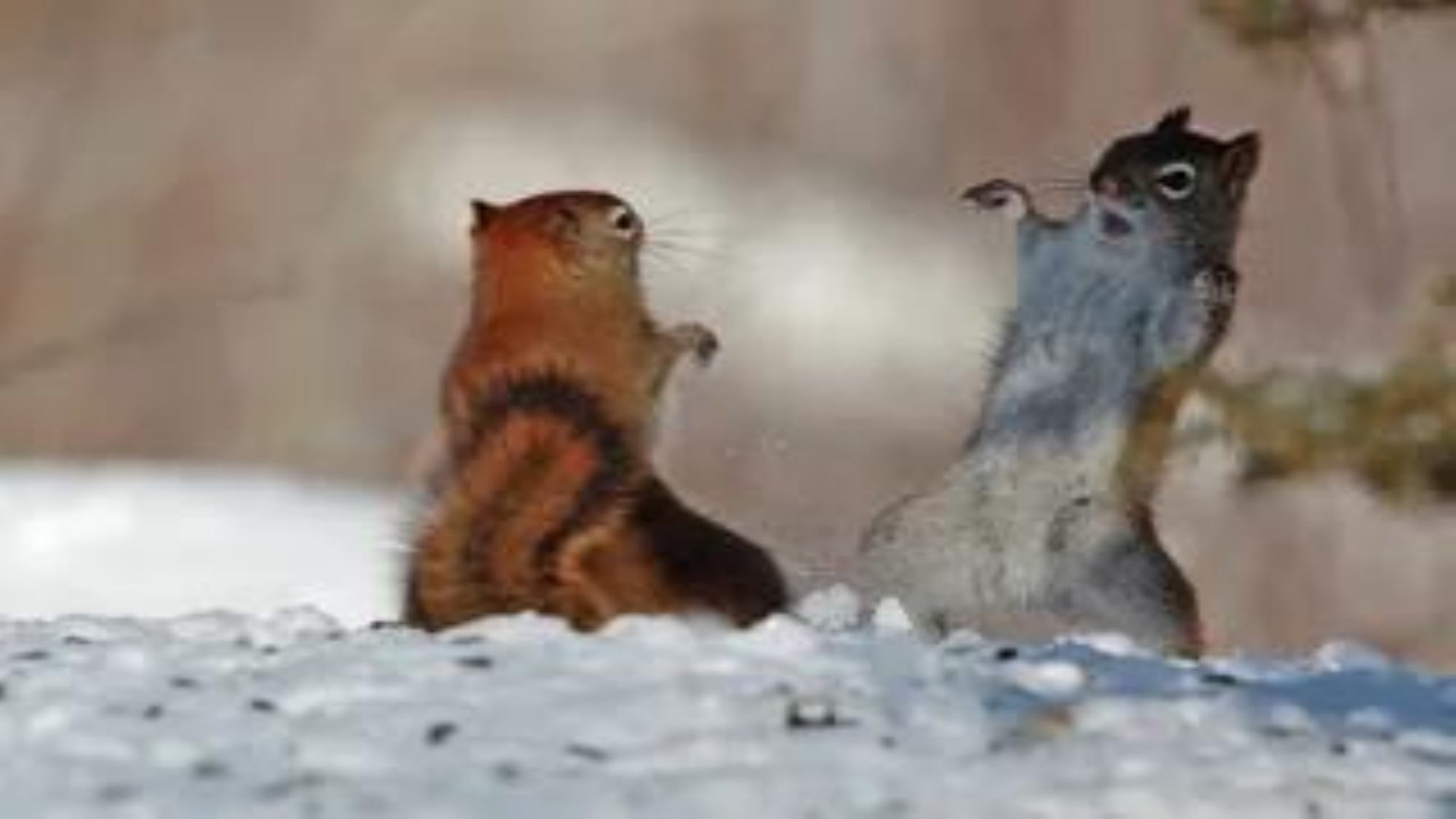}
    }
    \subfigure[\textbf{Caption:} Incredible new images of Jupiter from NASA. \newline  \textbf{Class: \color{red}\textbf{Misleading Content} } ]{
        \includegraphics[width=0.33\textwidth]{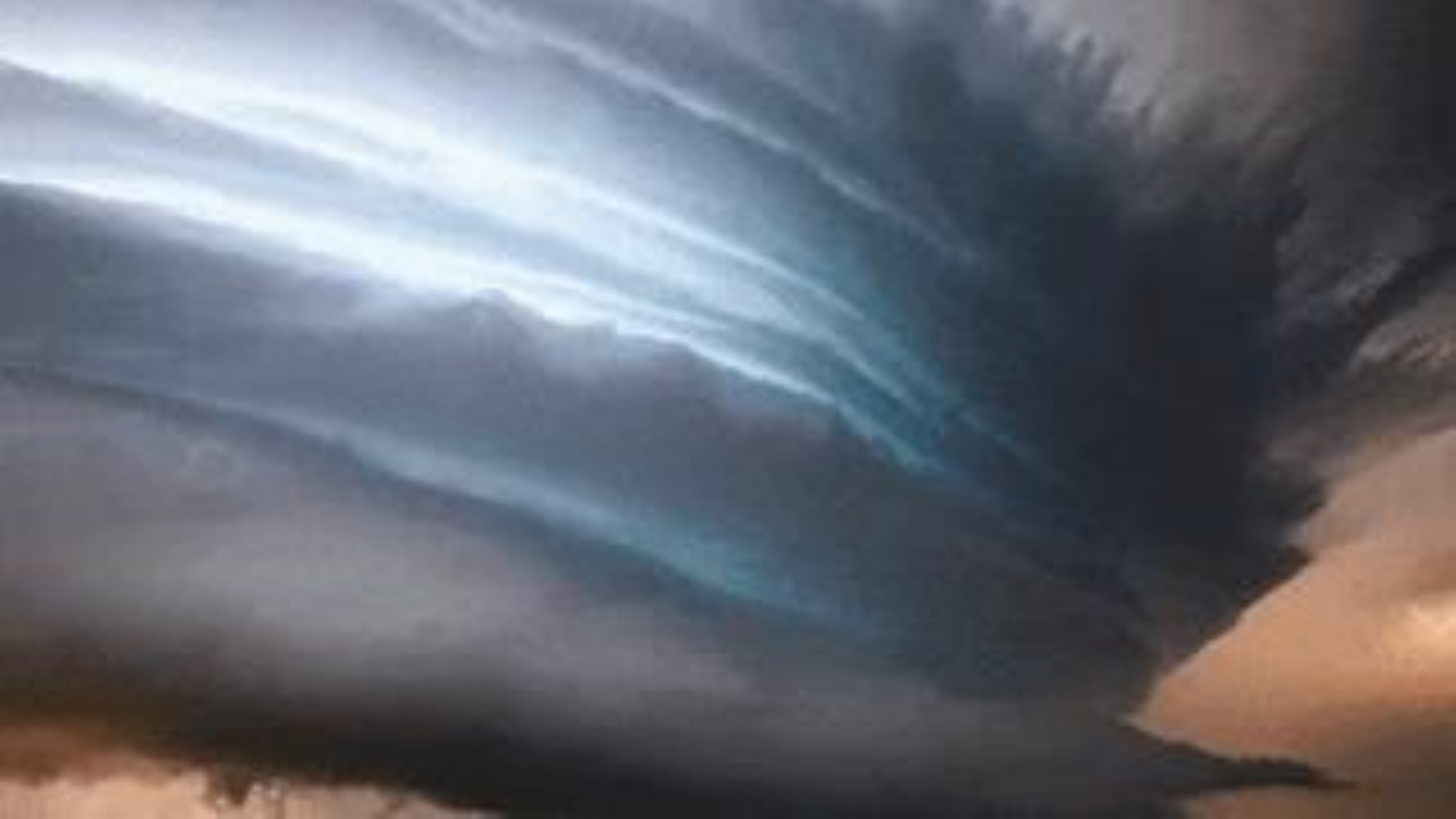}
    }
    \subfigure [\textbf{Caption:} Man binge-watches life passing him by. \textbf{Class: \color{red}\textbf{Satire/Parody} } ]{
        \includegraphics[width=0.33\textwidth]{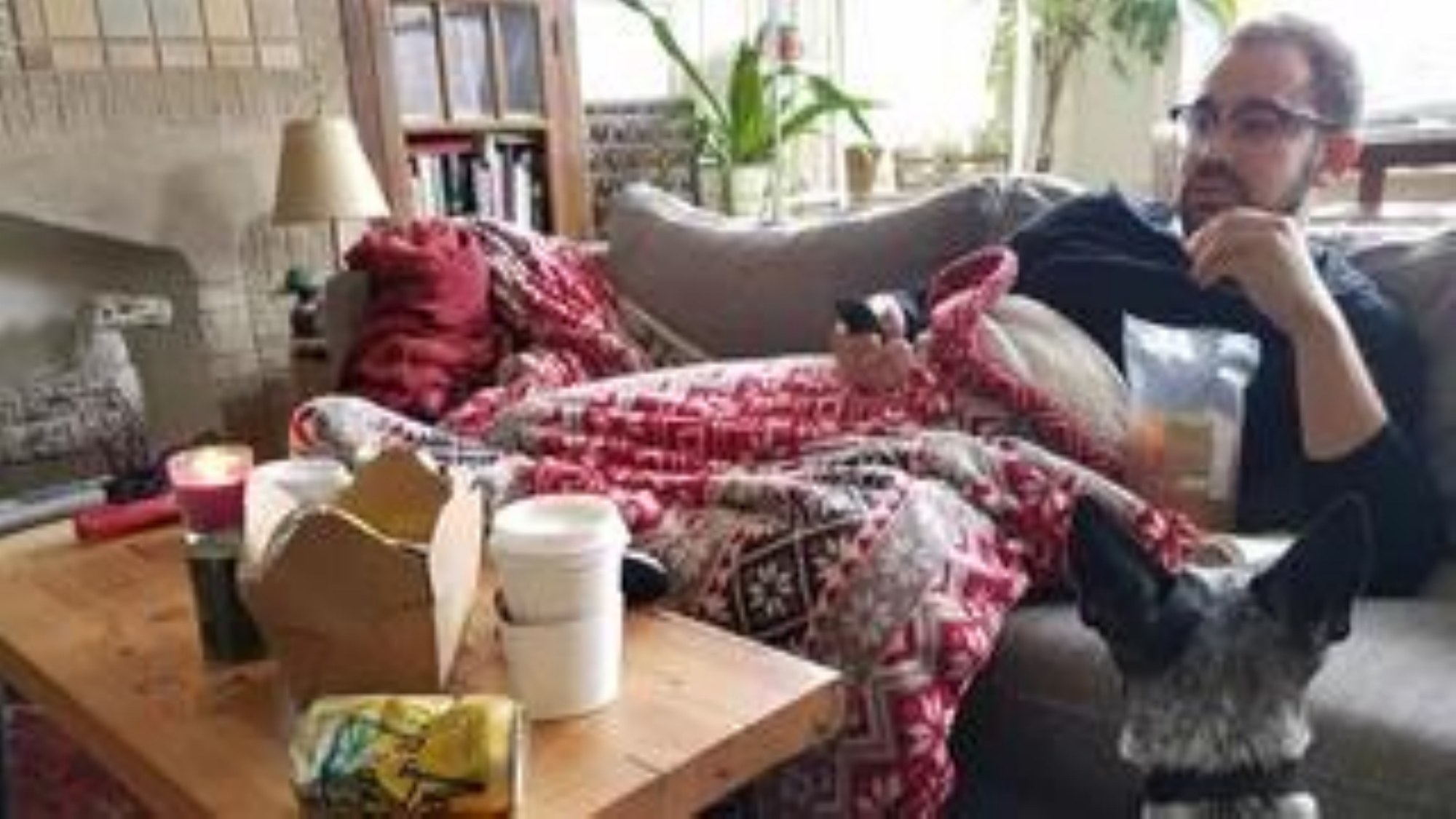}
    }
    \subfigure[\textbf{Caption:} King Arthur and his noble steed.\newline  \textbf{Class: \color{red}\textbf{Manipulated Content} }] {
        \includegraphics[width=0.33\textwidth]{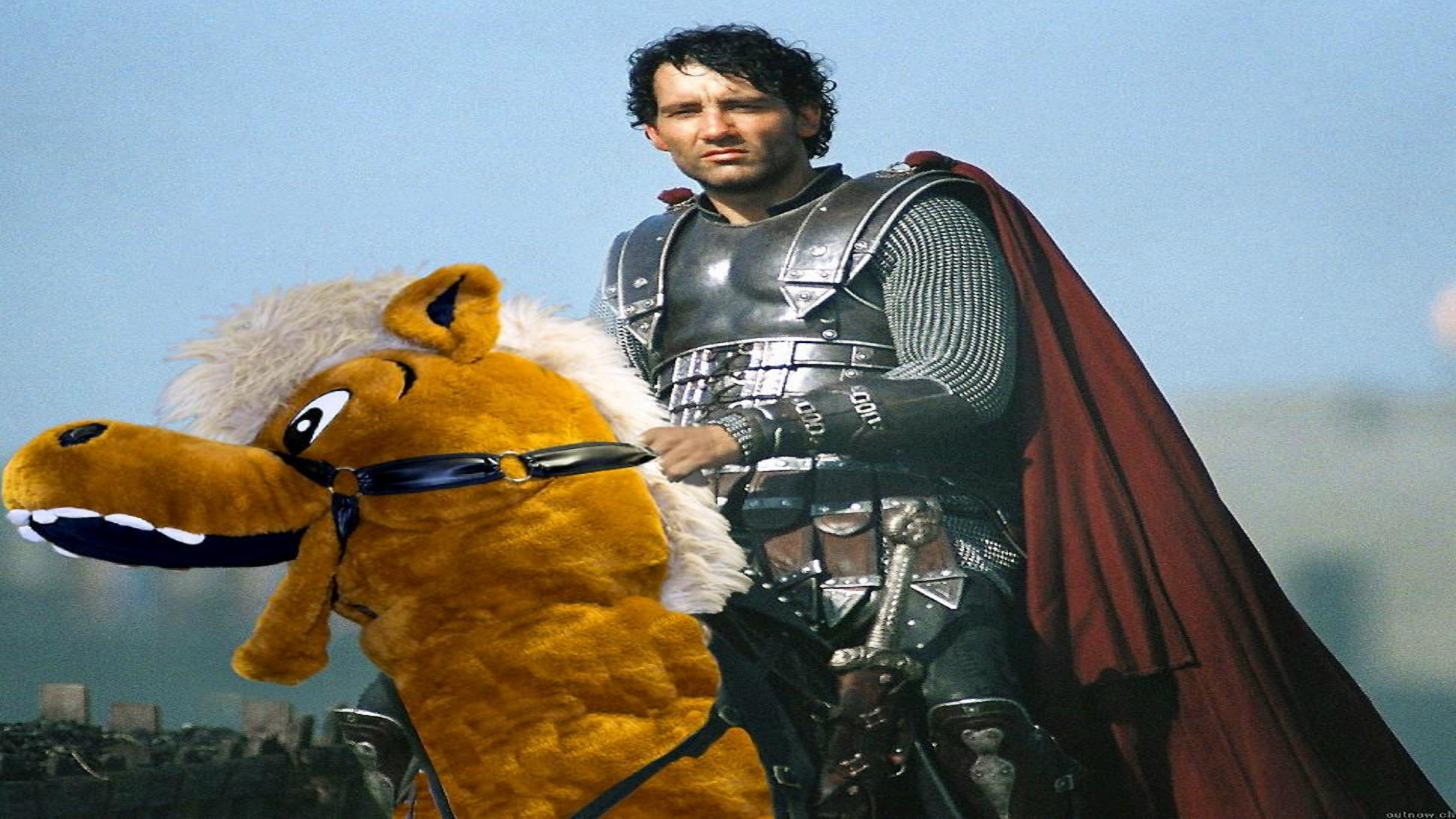}
    }
    \subfigure[\textbf{Caption: } Picture my sister left her corner after 30hrs, and had to go to first grade this year!\newline  \textbf{Class: \color{red}\textbf{False Connection} }]{
        \includegraphics[width=0.33\textwidth]{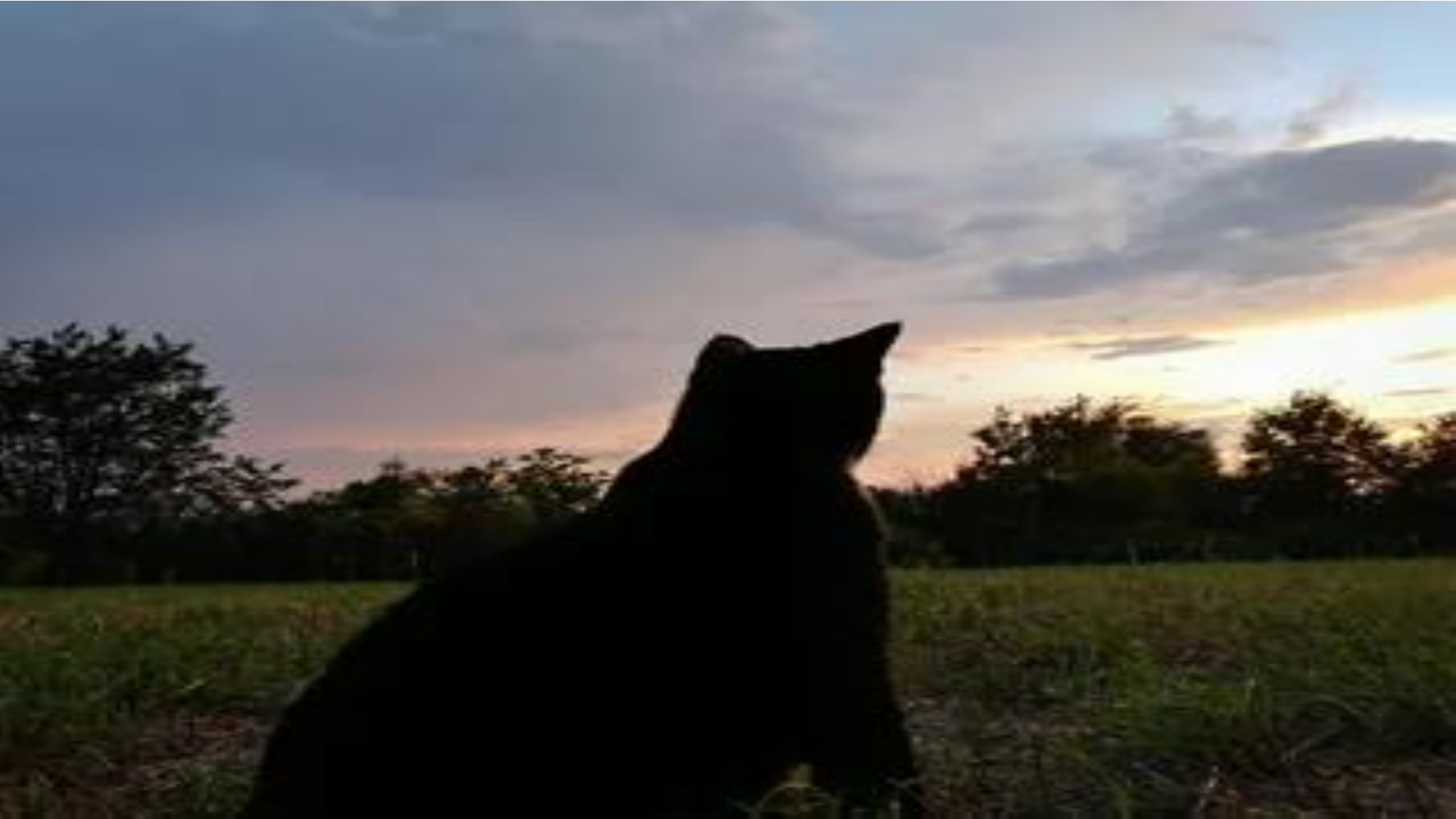}
    }
    \subfigure[\textbf{Caption:} ``One Raid by an Enemy Bomber...'' Leaflets dropped over US during Project Revere, an American Cold War propaganda test. \newline  \textbf{Class: \color{red}\textbf{Imposter Content} }]{
        \includegraphics[width=0.33\textwidth]{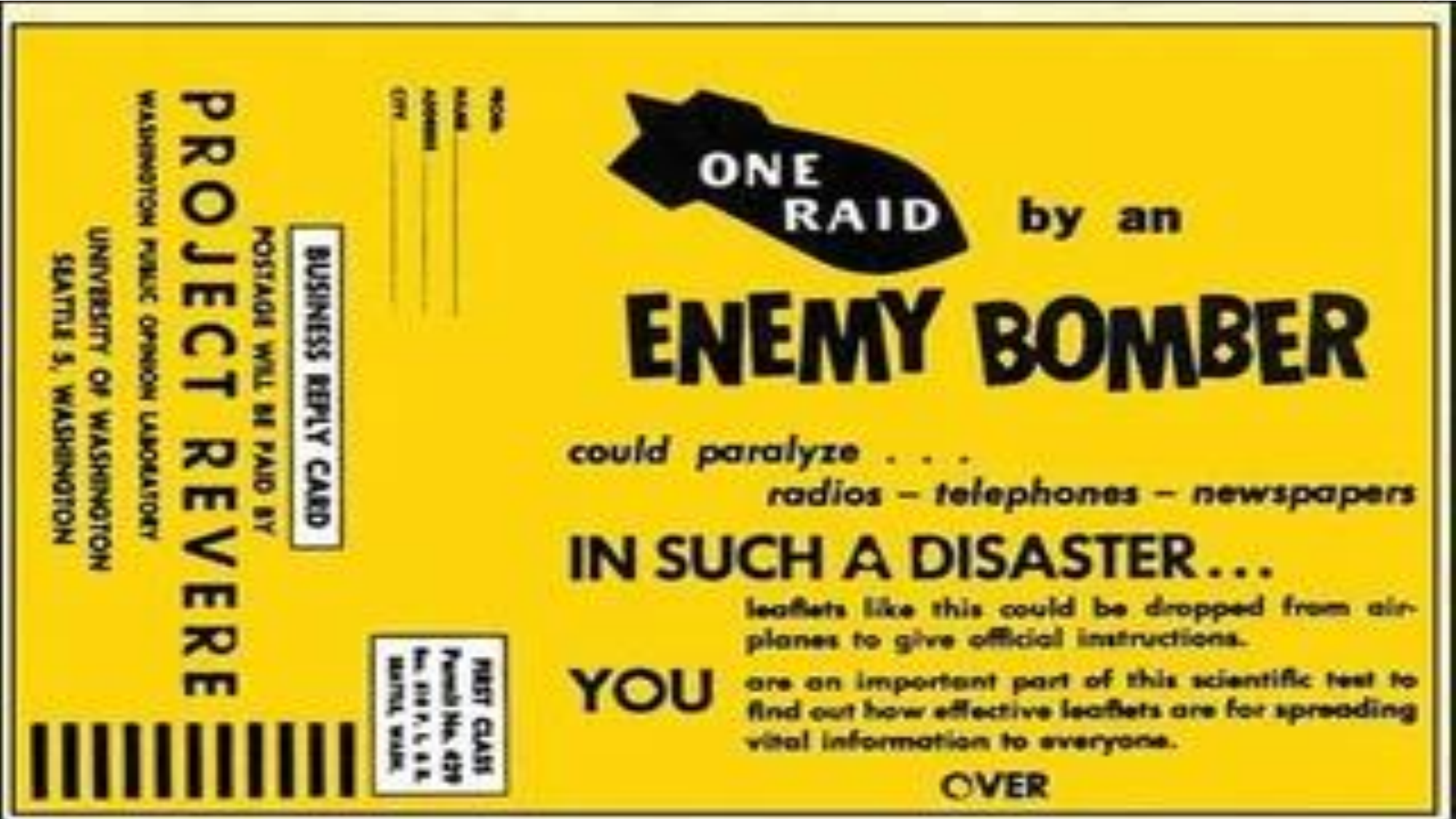}
    }
    \caption{Examples of different classes in \Fakeddit dataset~\cite{nakamura-etal-2020-fakeddit}.}
    \label{fig:subfigures}
\label{fig:fakeddit}
\end{figure}

\textbf{\Fakeddit:}\footnote{https://github.com/entitize/Fakeddit} is a dataset collected from Reddit, a social news and discussion website where users can post submissions on various subreddits. \Fakeddit consists of over 1 million submissions from 22 different subreddits spanning over a decade, with the earliest submission being from 3/19/2008 and the most recent submission being from 10/24/2019. These subreddits are posted on highly active and popular pages by over 300,000 users. \Fakeddit consists of submission titles, images, user comments, and submission metadata including score, the username of the author, subreddit source, sourced domain, number of comments, and up-vote to down-vote ratio. Approximately 64\% of the samples have both text and image data~\cite{nakamura-etal-2020-fakeddit}. Samples of this dataset are annotated with 2-way, 3-way, and 6-way labels including true, satire/parody, misleading content, manipulated content, false connection, and imposter content. Examples of 6-way labels are demonstrated in Fig.~\ref{fig:fakeddit}. Additionally, Table~\ref{table:comparison of methods} illustrates a comparison and evaluation of various methods' performance on the Fakeddit dataset~\cite{nakamura-etal-2020-fakeddit}\footnote{The table is based on the work in~\cite{nakamura-etal-2020-fakeddit}}.
\\
\begin{table}[t]
\centering
\caption{ Evaluation of classification accuracy on the \Fakeddit dataset using various image/text embedders conducted by~\cite{nakamura-etal-2020-fakeddit}.}
\label{my-label}
\small 
\begin{tabular}{@{}llp{1.5cm}p{1.5cm}lllllll@{}}
\hline
   &  &   & \multicolumn{2}{c}{2-way} & \multicolumn{2}{c}{3-way} & \multicolumn{2}{c}{6-way} \\ \hline
Type        & Text    & Image       & Validation    & Test       & Validation    & Test       & Validation    & Test       \\ 

\hline
Text+Image  & InferSent & VGG16    & 0.8655        & 0.8658     & 0.8618        & 0.8624     & 0.8130        & 0.8130     \\
            & InferSent & EfficientNet & 0.8328    & 0.8339     & 0.8259        & 0.8256     & 0.7266        & 0.7280     \\
            & InferSent & ResNet50 & 0.8888        & 0.8891     & 0.8855        & 0.8863     & 0.8546        & 0.8526     \\
            & BERT    & VGG16      & 0.8694        & 0.8699     & 0.8644        & 0.8655     & 0.8177        & 0.8208     \\
            & BERT    & EfficientNet & 0.8334      & 0.8318     & 0.8265        & 0.8255     & 0.7258        & 0.7272     \\
            & BERT    & ResNet50   & \textbf{0.8929} & \textbf{0.8909} & \textbf{0.8905} & \textbf{0.8900} & \textbf{0.8600} & \textbf{0.8588}     \\  \hline
\end{tabular}
\label{table:comparison of methods}
\end{table}

\textbf{\NewsBag}: comprises 200,000 real news and 15,000 fake articles. The real training articles have been collected from the Wall Street Journal and the fake ones from The Onion website\footnote{https://www.theonion.com/}, which publishes satirical content. However, the samples of the test set are collected from different websites, i.e., TheRealNews\footnote{https://therealnews.com/} and ThePoke\footnote{https://www.thepoke.co.uk/}. The rationale behind using different sources of news for the training and test sets is to observe how well the models could be generalized to unseen data samples. The \NewsBag dataset is highly imbalanced. Thus, to tackle this issue, \NewsBag++ is also released, which is the augmented training version of the \NewsBag dataset and contains 200,000 real and 389,000 fake news articles. Another weakness of the \NewsBag dataset is that it does not have any social context information such as spreader information, sharing trends, and reactions such as user comments and engagements~\cite{NewsBag_2020}.\\

\textbf{\MMCOVID}:\footnote{https://github.com/bigheiniu/MM-COVID} is a multi-lingual and multi-dimensional COVID-19 fake news data repository. This dataset comprises 3,981 fake news and 7,192 trustworthy information in 6 different languages, i.e., English, Spanish, Portuguese, Hindi, French, and Italian. \MMCOVID consists of visual, textual, and social context information, e.g., users and networks information~\cite{li2020mmcovid}. This dataset is annotated is by Snopes\footnote{www.snopes.com} and Poynter\footnote{www.poynter.org/coronavirusfactsalliance/} crowdsource domains where experts and journalists evaluate and fact-check news content and annotate contents as either fake or real. While Snopes is an independent publication that mainly contains English content, Poynter is an international fact-checking network (IFCN) which unites 96 different fact-checking agencies such as PolitiFact\footnote{https://www.politifact.com/} in 40 languages.\\

\textbf{\ReCOVery}:\footnote{https://github.com/apurvamulay/ReCOVery} contains 2,029 news articles that have been shared on social media, most of which (2,017 samples) have both textual and visual information for multi-modal studies. \ReCOVery is imbalanced in news class, i.e., the proportion of real vs. fake articles is around 2:1. The number of users who spread real news (78,659) and users sharing fake articles (17,323) is greater than the total number of users included in the dataset (93,761). In this dataset, the assumption is that users can engage in spreading both real and fake news articles. Samples of this dataset are annotated by two fact-checking resources: NewsGuard\footnote{https://www.newsguardtech.com/} and Media Bias/Fact Check (MBFC)\footnote{https://mediabiasfactcheck.com/}, which is a website that evaluates factual accuracy and political bias of news media. MBFC labels each news media as one of six factual-accuracy levels based on the fact-checking results of the previously published news articles. Samples of \ReCOVery are collected from 60 news domains, from which 22 are the sources of reliable news articles (e.g., National Public Radio\footnote{https://www.npr.org/} and Reuters\footnote{https://www.reuters.com}) and the remaining 38 are sources to collect unreliable news articles (e.g., Human Are Free\footnote{http://humansarefree.com/} and Natural News\footnote{https://www.naturalnews.com/})~\cite{CIKM'20_Zhou_recovery}.\\

\textbf{\CoAID}:\footnote{https://github.com/cuilimeng/CoAID} Covid-19 heAlthcare mIsinformation Dataset or \CoAID is a diverse COVID-19 healthcare misinformation dataset, including fake news on websites and social platforms, along with users’ social engagement about the news. It includes 5,216 news articles, 296,752 related user engagements, 926 social platform posts about COVID-19, and ground truth labels. The publishing dates of the collected information range from December 1, 2019, to September 1, 2020. In total, 204 fake news articles, 3,565 true news articles, 28 fake claims, and 454 true claims are collected. Real news articles are crawled from 9 reliable media outlets that have been cross-checked as reliable, e.g., National Institutes of Health (NIH)\footnote{https://www.nih.gov/news-events/news-releases} and CDC\footnote{https://www.cdc.gov/coronavirus/2019-ncov/whats-new-all.html}. Fake news is retrieved from several fact-checking websites, such as PolitiFact and Health Feedback\footnote{https://healthfeedback.org/}~\cite{cui2020coaid}.

\textbf{\MMCoVaR}: is a Multi-modal COVID-19 Vaccine Focused Data Repository (MMCoVaR). Articles in this dataset are annotated using two news website source checking methods, and the tweets are fact-checked based on a stance detection approach. \MMCoVaR comprises 2,593 articles issued by 80 publishers and shared between 02/16/2020 and 05/08/2021, and 24,184 Twitter posts collected between 04/17/2021 and 05/08/2021. Samples of this dataset are annotated by Media Bias Chart and Media Bias/Fact Check (MBFC) and classified into two levels of credibility: reliable and unreliable. Thus, articles are labeled as either credible or unreliable, and tweets are annotated as reliable, inconclusive, or unreliable~\cite{ASONOM'21_Chen_MMCoVaR}. It is worth mentioning that textual, visual, and social context information are available for the news articles.\\

\textbf{\NtfNews}:\footnote{https://github.com/billywzh717/N24News} is a multi-modal dataset extracted from the New York Times articles published from 2010 to 2020. Each news article belongs to one of 24 different categories, e.g., science, arts, etc. The dataset comprises up to 3,000 samples of real news for each category. In total, 60,000 news articles are collected. Each article sample contains a category tag, headline, abstract, article body, image, and corresponding image caption. This dataset is randomly split into training/validation/testing sets in the ratio of 8:1:1~\cite{wang2021n24news}. The main weakness of this dataset is that it does not have any fake samples, and all of the real samples are collected from a single source, i.e., The New York Times.\\

\textbf{\MuMiN}:\footnote{https://github.com/MuMiN-dataset/mumin-build}
Large-Scale Multilingual Multi-modal Fact-Checked Misinformation Social Network Dataset (MuMin) comprises 21 million tweets belonging to 26 thousand Twitter threads, each of which has been linked to 13 thousand fact-checked claims in 41 different languages. \MuMiN is available in three versions: large, medium, and small, with the largest one consisting of 10,920 articles and 6,573 images. In this dataset, if the claim is ``mostly true,'' it is labeled as factual. When the claim is deemed “half true” or ``half false,'' it is labeled as misinformation, with the justification that a statement containing a significant part of false information should be considered misleading content. When there is no clear verdict, the verdict is labeled as other~\cite{MuMiN}.

\par
\begin{figure*}[!h]
\centering
\includegraphics[width = 0.8\textwidth]{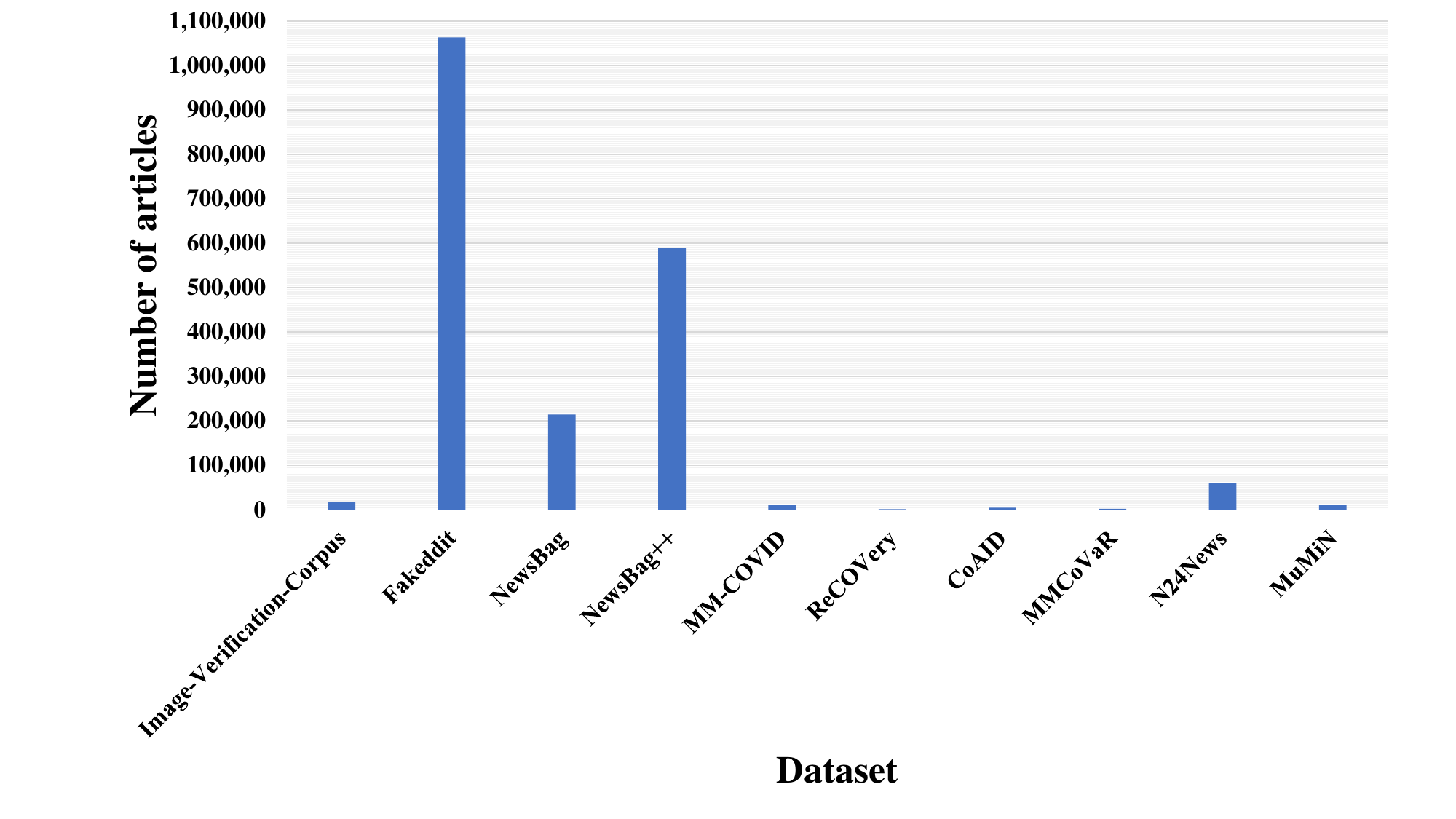}
\caption{Number of news articles by dataset.}
\label{fig:barchart}
\end{figure*}

 A summary and side-by-side comparison of the previously mentioned datasets are shown in Table \ref{table:tab1}. As illustrated in Fig. \ref{fig:barchart}, most of these datasets are small, annotated with binary labels, sourced from limited platforms like Twitter, and contain only a few modalities, namely text and image.

\renewcommand{\arraystretch}{1.6}

\begin{landscape}
\begin{center} 
\begin{table*}[!t]
\addtolength{\tabcolsep}{1pt}  
\small
\centering
\caption{Statistics of multi-modal databases for fake news detection}
 \begin{tabular}[lcr]{llllll}
 \toprule
\textbf{Dataset}&\textbf{Total Samples}& \textbf{\# classes}& \textbf{Modalities} & \textbf{Source}&\textbf{Details}\\
 \midrule
 \textbf{image-verification-corpus}~\cite{boididou2018detection}&17,806&2&image,text&Twitter\\
 
 \textbf{Fakeddit}~\cite{nakamura-etal-2020-fakeddit}&1,063,106&2,3,6&image,text&Reddit&\parbox{80pt}{682,996 samples are multi-modal}\\
 \textbf{NewsBag}~\cite{NewsBag_2020}&215,000&2&image, text&\parbox{70pt}{TTrain: Wall Street\&Onion.Test: TheRealNews \& ThePoke}&\parbox{110pt}{This dataset is highly imbalanced. There are only 15,000 fake samples.}\\\\
 \textbf{NewsBag++}~\cite{NewsBag_2020}&589,000&2&image,text&\parbox{70pt}{Train: Wall Street\&Onion. Test: TheRealNews \& ThePoke}&\parbox{110pt}{Same as NewsBag but fake samples are synthetic samples created by augmentation techniques}\\\\
 \textbf{MM-COVID}~\cite{li2020mmcovid}&11,173&2&\parbox{70pt}{image,text,social context}&Twitter&\parbox{110pt}{3,981 fake samples and 7,192 real samples}\\\\
 \textbf{ReCOVery}~\cite{CIKM'20_Zhou_recovery}&2,029&2&text,image&Twitter&\parbox{110pt}{ Imbalanced with ratio  of 2:1 real vs.fake}\\
  \textbf{CoAID}~\cite{cui2020coaid}&5,216&2&image,text&Twitter&\parbox{110pt}{Consists of 296,752 user engagements (926 social platforms)}\\
 \textbf{MMCoVaR}~\cite{ASONOM'21_Chen_MMCoVaR}&\parbox{75pt}{2,593 articles \& 24,184 tweets}&2&\parbox{70pt}{image,text,social context}&Twitter&\parbox{110pt}{Tweets as labeled as reliable,inconclusive and unreliable}\\
 \textbf{N24News}~\cite{wang2021n24news}&60,000&24&image,text&New York Times& \parbox{110pt}{All samples are real from 24 different categories}\\
 \textbf{MuMiN}~\cite{MuMiN}&10,920&3&image,text&Twitter&\parbox{110pt}{ Consists of 10,920 articles and 6,573 images.}\\
 \bottomrule
\end{tabular}
\label{table:tab1}
\end{table*}
\end{center}
\end{landscape}

\section{CHALLENGES IN MULTI-MODAL MISINFORMATION
DETECTION}

\label{sec:challenges}
Recent studies on multi-modal learning have made significant contributions to the field of multi-modal fake news detection. However, there are still weaknesses and shortcomings, and recognizing
them opens the door to new opportunities not only in fake news detection but also in the multi-modal field in general. In this section, we discuss challenges, shortcomings, and opportunities in
multi-modal fake news detection. We provide non-exhaustive lists of challenges and shortcomings
for each direction of multi-modal misinformation research.

\subsection{Data study Challenges}
This category refers to the weaknesses of current multi-modal datasets for misinformation detection.
We briefly discussed some of these weaknesses in the multi-modal data study section. An itemized
list of such limitations and shortcomings is as follows:
\begin{itemize}
\item  \textbf {Lack of large and comprehensive datasets:} as illustrated in Fig.~\ref{fig:barchart}, most of the existing
datasets are small in size and sometimes highly imbalanced in terms of the fake-to-real ratio.
\item  \textbf { Lack of cross-lingual datasets:} almost all social media platforms are multi-lingual environments where users share information in multiple languages. Although misinformation
spreads in multiple languages, a vast majority of the existing datasets are mono-lingual, i.e.,
they only provide English content. Therefore, there is a serious lack of non-English content
and annotations.
\item  \textbf {Limited modalities:} as we discussed earlier, most of the existing multi-modal datasets only
provide image and text modalities, thus neglecting useful information conveyed by other
modalities such as video, audio, etc. The necessity of providing more modalities becomes
more apparent when we consider popular social media such as YouTube, TikTok, and
Clubhouse, which are mainly video or audio-based platforms.
\item  \textbf { Bias in event-specific datasets:} many of the existing datasets are created for specific events
such as the COVID-19 crisis, thereby not covering a variety of events and topics. As a result,
they may not sufficiently train models to detect fake news in other contexts.
\item  \textbf {Binary and domain-level ground truth:} most of the existing datasets provide binary
and domain-level ground truth for well-known outlets such as The Onion or The New
York Times. In addition, they often do not provide any information about the reasons for
misinformation, e.g., cross-modal discordance, false connection, imposter content, etc.
\item  \textbf {Subjective annotations and inconsistency of labels:} as discussed in the data study section,
different datasets use different crowd-sourced and fact-checking agencies, thereby articles are annotated subjectively with different labels across different datasets. Thus, it is very
challenging to analyze, compare, and interpret results.
\end{itemize}

\subsection{Feature Study Challenges} This category comprises shortcomings related to cross-modal feature identification and extraction in the multi-modal fake news detection pipeline. Some of the most important weaknesses in current feature-based studies are:

\begin{itemize}
\item{\textbf{Insufficiency of cross-modal cues:}} although researchers have proposed some multi-modal cues, most of the existing models naively fuse image-based features with textual features as a supplement. There are fewer works that leverage explainable cross-modal cues other than image and text combinations. However, there are still plenty of useful multi-modal cues that are often neglected by researchers.
\item{\textbf{Ineffective cross-modal embeddings:}} as mentioned earlier, the majority of the existing approaches only fuse embeddings with simple operations such as concatenation of the representations, thereby failing to build an effective and non-noisy cross-modal embedding. Such architectures fail in many cases, as the resulting cross-modal embedding consists of useless or irrelevant parts which may result in noisy representations.

\item{\textbf{Lack of language-independent features:}} a majority of existing work on misinformation leverages text features that are highly dependent on dataset languages, which are mostly English. Identifying language-independent features is an effective way to cope with mono-lingual datasets.
\end{itemize}

\subsection{Model Study Challenges} This category refers to the shortcomings of current machine learning solutions in detecting misinformation in multi-modal environments. The following is a non-exhaustive list of existing shortcomings:
\begin{itemize}

   \item{\textbf{Inexplicability of current models:}} a majority of the existing models do not provide any explicable information about the regions of interest, common patterns of inconsistencies among modalities, and types of misinformation (e.g., manipulation, exaggeration). While some recent works attempt to use attention-based techniques to overcome the problem of ineffective multi-modal embedding and provide some interpretability, most of them usually follow a trial-and-error approach like masking to find relevant sections to attend to. However, interpretable and explainable AI is crucial in building trust and confidence as well as fairness and transparency, which are mostly neglected.
   
  \item{\textbf{Non-transferable models to unseen events:}} most of the existing models are designed in such a way that they extract and learn event-specific features (e.g., COVID-19, election). Thus, they are most likely biased toward specific events and, as a result, not transferable to unseen and emerging events. For this reason, building models that learn general features and separate them from the non-transferable event-specific features would be extremely useful.
   
   \item{\textbf{Unscalability of current models:} }considering the expensive and complicated structures of deep networks and the fact that most of the existing multi-modal models leverage multiple deep networks (one for each modality), they are not scalable if the number of modalities increases. Moreover, many of the existing models require heavy computing resources and need a large volume of memory storage and processing units. Therefore, the scalability of proposed models should be taken into account while developing new architectures.
   
\item{\textbf{Vulnerabilities against adversarial attacks:}} malicious adversaries continuously try to fool the misinformation detection models. This is especially feasible when the underlying model’s techniques and cues are revealed to the attacker, such as when the attacker can probe the model. As a result, many of the detection techniques become dated in a short period of time. Thus, there is a need to create detection models that are resistant to manipulation.
\end{itemize}

\section{Opportunities in Multi-modal Misinformation Detection} 
\label{sec:opportunities}
Considering the challenges and shortcomings in multi-modal misinformation detection we discussed above, we propose opportunities for furthering research in this field. In what follows, we discuss these opportunities by each direction of multi-modal misinformation detection study.

\subsection{Opportunities in Multi-modal Data Study} Considering the data study challenges we discussed earlier, we propose the following avenues:
\begin{itemize}
\item{\textbf{Comprehensive multi-modal and multi-lingual datasets:}} as we discussed earlier, one important gap in the misinformation detection study is the lack of a comprehensive multi-modal dataset, which needs to be addressed in the future. Multi-modal misinformation detection requires large, multi-lingual, multi-source datasets that cover a variety of modalities, web resources, events, etc., and provide fine-grained ground truth for the samples.

\item{\textbf{Standardized annotation strategy:}} Current datasets are annotated by various fact-checking agencies, leading to subjective labels in many cases. Establishing a standardized labeling agreement across all datasets would facilitate easier cross-dataset comparison and analysis.
\end{itemize}
\subsection{Opportunities in Multi-modal Feature Study} Based on the feature study challenges we discussed in the previous section, we propose the following research opportunities to overcome some of the existing challenges in multi-modal feature study: 
\begin{itemize}
\item{\textbf{Identifying cross-modal clues:}} Currently, cross-modal cues are restricted to a few basic indicators, such as the similarity between text and images. Identifying more subtle and often overlooked cues can aid in developing discordance-aware models and help recognize vulnerabilities in the serving platforms, which is integral to adversarial learning.
\item{\textbf{Developing efficient fusion mechanisms:}}many of the existing solutions leverage naive fusion mechanisms such as concatenation, which may result in inefficient and noisy multi-modal representations. Therefore, another fruitful avenue of research lies in the study and development of more efficient fusion techniques to produce information-rich representations.

\item{\textbf{Identifying language-independent features to cope with mono-lingual datasets:}} a majority of existing datasets are mono-lingual, thereby not sufficient enough to train models for non-English tasks. One way to compensate for the lack of multi-lingual datasets is to use language-independent features~\cite{Vogel2020DetectingFN}. Identifying such features, especially in multi-modal environments where there are more features and aspects, would be highly effective in coping with mono-lingual datasets.
\end{itemize}
\subsection{Opportunities in Multi-modal Model Study} Some unexplored research avenues to tackle existing model-related challenges in multi-modal misinformation detection include: 
\begin{itemize}

\item{\textbf{Utilizing foundation models and prompt-based techniques in multi-modal misinformation detection:} }The astounding effectiveness of foundation models and techniques, including ICL and prompt-tuning, in numerous multi-modal tasks suggests that foundation models have a lot of potential for identifying multi-modal misinformation. Developing task-specific foundation models for detecting misinformation is another opportunity that would hugely impact the field of misinformation detection.
\item{\textbf{Developing cross-modal discordance-aware architectures:}} most of the existing works either blindly merge modalities or take a trial-and-error approach to attend to the relevant modalities. Implementing discordance-aware models not only results in information-rich representations but also may be useful in making attention-based techniques more efficient.

\item{\textbf{Adversarial learning in multi-modal misinformation detection:}} Although there are existing generative-based architectures, adversarial study of multi-modal misinformation detection has been mostly neglected. To make the detection models more adversarially robust, it is of utmost importance to dedicate time and effort to the study and development of generative and adversarial learning techniques.

\item{\textbf{Interpretability of multi-modal models:}} Development of explainable frameworks to help better understand and interpret predictions made by multi-modal detection models is another opportunity in multi-modal misinformation detection. Explicability can be very useful for related tasks such as the predictability of models, fairness and bias, and adversarial learning.

\item{\textbf{Transferable models to unseen events:}} As mentioned earlier, except for a few works, most of the existing models are designed for specific events and, as a result, are ineffective for emerging ones. Since misinformation spreads during a variety of events, developing general and transferable models is extremely crucial.

\item{\textbf{Development of scalable models:}}Another opportunity is to develop models that are more efficient in terms of time and resources and do not become intolerably complicated while increasing the number of fused modalities.
\end{itemize}

\section{Conclusions}
In this paper, we review the literature on multi-modal misinformation detection, discuss its strengths and weaknesses, and suggest new directions for future research. First, we introduce some of the prominent misinformation categories and often-used cross-modal cues for spotting them. We also discuss different fusion mechanisms to merge modalities that are engaged in such cross-modal cues. In addition, we categorize existing solutions into two groups: classic machine learning and deep learning solutions, and then further divide each group based on the techniques that are utilized. Furthermore, we introduce and compare existing datasets on multi-modal misinformation detection and identify some of the weaknesses of these datasets. By classifying them into data, feature, and model-based shortcomings, we demonstrate some of the most prominent problems in multi-modal fake news detection. Finally, we propose new lines of research to address these shortcomings.
\newpage


\end{document}